\documentclass[10pt,journal,compsoc]{IEEEtran}

\normalsize

\usepackage[cmex10]{amsmath}
\usepackage{graphicx}
\usepackage{epstopdf}
\usepackage{multirow}
\usepackage{amssymb}
\usepackage{algorithm}
\usepackage[noend]{algpseudocode}
\usepackage{tensor}
\usepackage{boondox-cal}
\usepackage{ifthen,boxedminipage}
\usepackage{cite}
\usepackage{lipsum}
\usepackage{mathtools}
\usepackage{cuted}
\usepackage{varwidth}
\usepackage{tabularx}
\usepackage{rotating}
\usepackage{times,color}

\DeclareMathOperator{\csch}{csch}
\DeclareMathOperator{\Lip}{Lip}
\DeclareMathOperator{\Vol}{vol}
\definecolor{dgreen}{rgb}{0,0.48,0.3}

\begin{document}

\title{Improving embedding of graphs with missing data by soft manifolds}

\author{Andrea Marinoni, Pietro Li\`{o}, Alessandro Barp, Christian Jutten, Mark Girolami
	\thanks{A. Marinoni is with Dept. of Physics and Technology, UiT the Arctic University of Norway, P.O. box 6050 Langnes, NO-9037, Tromsø, Norway. 
	E-mail: andrea.marinoni@uit.no.
	P. Li\`{o} is with Dept. of Computer Science and Technology, University of Cambridge, 15 JJ Thomson Avenue, Cambridge CB3 0FD, UK. E-mail: pl219@cam.ac.uk.
	A. Barp is with The Alan Turing Institute, 96 Euston Rd, London NW1 2DB, UK. E-mail: abarp@turing.ac.uk. 
	C. Jutten is with GIPSA-lab, University of Grenoble Alpes, 38000 Grenoble, France, and with the Institut Universitaire de France, 75005 Paris, France. E-mail: christian.jutten@gipsa-lab.grenoble-inp.fr. 
	M. Girolami is with Dept. of Engineering, University of Cambridge, Trumpington St., Cambridge CB2 1PZ, UK, and with The Alan Turing Institute, 96 Euston Rd, London NW1 2DB, UK. E-mail: mag92@cam.ac.uk. 
	}
}

\IEEEtitleabstractindextext{%
\begin{abstract}
Embedding graphs in continous spaces is a key factor in designing and developing algorithms for automatic information extraction to be applied in diverse tasks (e.g., learning, inferring, predicting). 
The reliability of  graph embeddings directly depends on how much the geometry of the continuous space matches the graph structure. 
Manifolds are mathematical structure that can enable to incorporate in their topological spaces the graph characteristics, and in particular nodes distances. 
State-of-the-art of manifold-based graph embedding algorithms take advantage of the assumption that the projection on a tangential space of each point in the manifold (corresponding to a node in the graph) would locally resemble a Euclidean space. 
Although this condition helps in achieving efficient analytical solutions to the embedding problem, it does not represent an adequate set-up to work with modern real life graphs, that are characterized by weighted connections across nodes often computed over sparse datasets with missing records. 
In this work, we introduce a new class of manifold, named soft manifold, that can solve this situation. 
In particular, soft manifolds are mathematical structures with spherical symmetry where the tangent spaces to each point are hypocycloids whose shape is defined according to the velocity of information propagation across the data points. 
Using soft manifolds for graph embedding, we can provide continuous spaces to pursue any task in data analysis over complex datasets. 
Experimental results on reconstruction tasks
on synthetic and real datasets show how the proposed approach enable more accurate and reliable characterization of graphs in continuous spaces with respect to the state-of-the-art. 
	

\end{abstract}

}

\maketitle

\section{Introduction} 
\label{sec_intro}

Recently, graph-based data analysis has gained strong interest by the scientific community for the investigation of complex datasets. 
This resulted by the proven ability of graph to provide a compact structure that can be used to effectively and efficiently explore the properties of the considered data \cite{GEOMDL,fluid_graphsignalproc}. 
In particular, the samples in the considered dataset are typically represented as nodes in the graph structure, whilst the edges are meant to model and quantify the similarity between each pair of nodes, according to a given informativity criterion and/or distance metric \cite{fluid_graph_learn_smooth,fluid_graphlearn_DF,fluid_graphlearn_SP,fluid_graphsignalproc,CommunityDetection_Fortunato,GEOMDL}. 
In this way, it is possible to obtain a thorough understanding of the interactions among samples. 
This in turn enables to achieve a solid characterization of the analysis to multiple tasks and across various domains.

A full understanding of the processes underlying the graph definition can be achieved by means of representation learning, which relies on identifying and defining a continuous space that would embed the characteristics of the datasets under exam \cite{HeterManifold,Fluid_multimodML,fluid_arxiv,machlearPR3, ReprLearn1,ReprLearn2,ReprLearn3,ReprLearn4, ReprLearn5, ReprLearn6, ReprLearn7}. 
This step is then typically followed by processing operations aiming at extracting information at different levels and for diverse purposes, e.g., learning, inferring, transferring, assessing, and predicting. 
\begin{figure*}[htb]
	\centering
	\includegraphics[width=2\columnwidth]{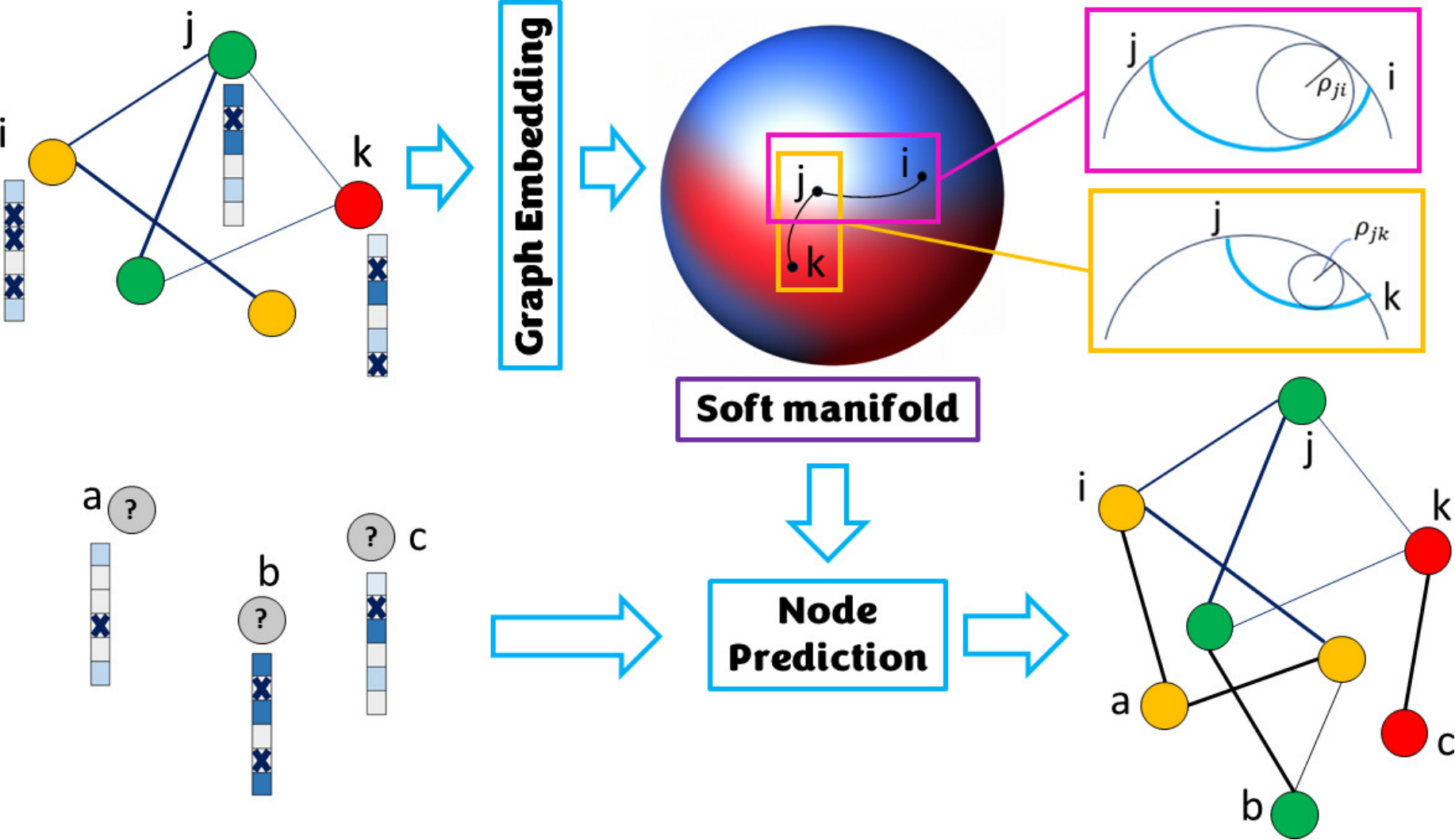} 
	\caption{Graphical abstract summarizing the main steps of this work. We propose a new method to perform embedding of a graph onto a new type of manifold (soft manifold) in continuous space.  The nodes in this graph (\textit{i}, \textit{j}, and \textit{k} in this Figure) are characterized by a set of features (identified by the arrays in Figure, where the different colors represent different values of the corresponding features). Some of these features might be missing (crosses in these arrays). The soft manifold show spherical symmetry and is considered a realization of the fluid diffusion mechanism. The colors on the soft manifold surface model the mechanical characteristics of the given dataset (e.g., conductivity, diffusivity) retrieved by the corresponding fluid dynamics modelling. This ultimately leads to different distances (computed by means of hypocycloid functions) between data points (see the sections of the hyperplane orthogonal to the surface of the soft manifold in magenta and orange boxes on the top right of the Figure). Once the properties of the soft manifold are learnt by means of graph embedding, it is possible to infer the labels to be associated with the nodes that were not used in graph embedding (identified by nodes \textit{a}, \textit{b} and \textit{c} in this Figure - note that their labels were not known before node prediction, and they were assigned a '?' label): the labels are identified by the color of the nodes in the graph - see bottom right in the Figure. }
	\label{fig_graphAbstr}
\end{figure*}

Encoding the graph characteristics on complex and heterogeneous manifolds is a key component of graph-based representation learning \cite{HeterManifold,GraphEmbed1,GraphEmbed2,GraphEmbed3}. 
State-of-the-art methods aim to match both pairwise distances and node-wise curvature information with pointwise curvature on the manifold. 
In this way, the resulting manifold are able to summarize the configuration of the embedded nodes, as well as the graph structural properties and conditions. 
Thus, the manifolds are ultimately able to display both distance and curvature information in building graph representations. 

These recent manifold learning strategies have been employed rather successfully in diverse research fields.  
However, their capacity to address operational scenarios is still very much limited. 
In fact, 
%
%
modern data analysis is more and more characterized by the investigation of datasets where several features for each data sample might not be relevant (e.g., being corrupted, or resulting by linear correlation of combinations of features), or even missing, e.g., because of the diversity in resolutions and acquisition strategies used by the various sensors used to conduct the analysis. 
Since state-of-the-art methods for graph embedding require smoothness and differentiability of the manifolds, hence all the records associated with each data point are used to compute the distances among samples. 
Thus, in these conditions state-of-the-art schemes for graph embedding would require to fill the data gaps with classic data processing techniques (e.g., impainting, interpolation). 
The validity of the analysis outcomes is then jeopardized, since inferring the data values can essentially oversimplify the structures of the data, hence not allowing us to capture the intrinsic complexity of the underlying phenomena and processes. 

To address these problems, we focused our attention on the design of a new type of manifolds that would enable to perform graph embedding also in case of missing data without any artificial inputs resulting by operations such as impainting or interpolation. 
This new mathematical structure, named \textit{soft manifolds}, rely on a different characterization of the manifold topology and local geometry. 
Specifically, whilst all manifolds in technical literature assume that each data point on the manifold could be locally inscribed in a tangent space that resembles a Euclidean hyperplane, soft manifolds assume that each tangent space is not Euclidean flat anymore. Thus, the distance between points can be computed as a hypocycloid whose properties depend on the velocity of information propagation across data points. 

By taking advantage of this novel manifold representation, we develop a new graph embedding algorithm that can be analytically described and easily implemented in a convex optimization scheme. 
This allows us to retrieve accurate and reliable information about the continuous space induced by the graphs in several operational scenarios with high efficiency. 
In particular, in this work we focused our attention to node prediction task. The experimental tests we conducted on diverse datasets affected by missing data show the robustness and the precision of the proposed approach, and support the actual improvement of the proposed strategy over state-of-the-art methods. Figure \ref{fig_graphAbstr} summarizes these points.

The main contributions of this paper can be summarized as follows: 
\begin{itemize}
	\item a novel type of manifold able to adapt to the description of complex datasets affected by sparsity and/or missing data is introduced; 
	\item a characterization of manifold topology in terms of fluid diffusion-based information propagation;
	\item a novel algorithm for weighted graph embedding based on a novel description of long-range interactions among data samples and local geometrical features is proposed, so that a thorough and interpretable understanding of the encoded data geometry is possible;
\end{itemize}


The paper is organized as follows. 
Section \ref{sec_back} reports the state-of-the-art of graph embedding on manifolds, the main limitations of the algorithms that are currently employed for this task, as well as the motivations for this work. 
In Section \ref{sec_meth_all}, we first briefly summarize the main elements of the fluid diffusion model that is used to derive the graph representation introduced in \cite{fluid_arxiv}. 
Then, we introduce the principles and properties of the soft manifolds, and their  characteristics in terms of invariance, mapping, metric and distances between points are clearly outlined. 
Finally, we provide the main steps for the graph embedding algorithm based on soft manifold representation. 
Section \ref{sec_result} reports the performance results obtained over two real life datasets with missing data, showing the actual impact of the proposed algorithm for graph embedding with respect to state-of-the-art schemes. 
Finally, Section \ref{secconcl} delivers our final remarks and some ideas on future research.
%

\section{Background and motivations}
\label{sec_back}

The topic of manifold learning and graph embedding has been widely investigated in technical literature at theoretical, methodological, and operational levels \cite{HeterManifold,GraphEmbed1,GraphEmbed2,GraphEmbed3,ReprLearn1,ReprLearn3,ReprLearn4,GEOMDL,GEOMDL_limits_1,fluid_graphsignalproc,GraphEmbedTheory1,GraphEmbedTheory2}. 
In particular, the development of graph neural networks has supported the exploration of non-Euclidean geometries for representation learning and graph embedding on continuous spaces \cite{GEOMDL,fluid_graphsignalproc,fluid_structeqmodel2,fluid_structeqmodel1,GraphEmbed3, GraphEmbed4,GraphEmbed5,GraphEmbed6,GraphEmbed7}. 
Also, non-Euclidean geometry has been used to investigate undesired effects in data analysis, hence boosting the characterization of geometrical features and properties of the considered set of records \cite{GEOMDL,GraphEmbed7,GraphEmbed5,GraphEmbed4}. 
Also, hyperbolic embeddings have been successfully used for graph reconstruction operations and description of interactions in natural systems \cite{GraphEmbed11,GraphEmbed10,GraphEmbed12,GraphEmbed9,GraphEmbed14,GraphEmbed15}. 

Exploiting the properties of hyperbolic geometries have played a key role for graph embedding. 
Specifically, it has been proven how the structural characteristics of the graphs can be efficiently encoded in hyperbolic spaces, especially by adapting the curvature of the manifolds \cite{GraphEmbed15,GraphEmbed14,GraphEmbed13,GraphEmbed12,GraphEmbed11,HeterManifold}. 
This approach provides higher flexibility in tracking the graph structural properties than classic Euclidean settings, and has been explored by means of diverse strategies, e.g., focusing on the product of curvature spaces, or analyzing matrix manifolds. 

In particular, 
allowing the manifolds to show non constant curvature has been proven a good choice to improve the matching between manifold and graph structures \cite{GraphEmbed2,GraphEmbed3,GraphEmbed1}. 
To guarantee efficiency in the embedding optimization, assuming spherical symmetry on the manifold topology has been identified as a key step \cite{HeterManifold,GraphEmbed10}. 
This in fact enables an analytical definition of the distances and projection maps that are in turn used to employ simple optimization schemes (such as gradient descent) to perform the graph embedding. 
The resulting heterogeneous spherical manifolds are able to address the variability in the graph structural properties, e.g., density of the nodes, nodes with variable degrees \cite{HeterManifold}. 

Although these state-of-the-art strategies are able to accommodate very important features for the use of graph embedding schemes in applied scenarios, it is also true that they cannot address a number of urgent problems of modern data analysis. 
This is particularly true when considering operational use of data analysis architectures. 
In fact, the technological advancements that have taken place in the last decades in sensing and monitoring has enabled the emergence of a huge variety of sensors and measurement stations that can be used to describe and characterise every event in our everyday life. This represents an incredible opportunity to improve the understanding of the interaction between human and environment, and consequently enhance human welfare. 
All basic sectors in society (industry, government, research, environment and civil society) can take advantage of this, and data analysis has the opportunity to improve the quality of decisions and planning in both private and public sectors \cite{Tuia_survey,Operational1,Persello_survey,Operational2}.These opportunities can be sparked by processing data that:

\begin{itemize}
\item \textbf{C1}: are massive, as the advancements in sensing and monitoring allow the storage of huge loads of records;
\item \textbf{C2}: show multiple resolutions (either in time, space, metrical units);
\item \textbf{C3}: are acquired at irregular time intervals, depending on the sensing technology (e.g., a satellite can orbit over one given region every two weeks, while a ground measurement station can acquire data every minute over the same region).
\end{itemize}

These conditions lead to the emergence of datasets characterized by missing and corrupted data, unevenly distributed across the considered data points (i.e., one data point might be characterized by $n_1$ features, whilst another one might be characterized by $n_2 \neq n_1$ features). 
Addressing this sort of situation makes all the strategies for identification of continuous spaces for embedding that have been introduced in technical literature unable to guarantee the extraction of accurate and reliable information. 

Specifically, using state-of-the-art manifold representation to address missing data situations implies leveraging some of the characteristics of the problem at hand (e.g., using the same subset of features to analyse subsets – or even entire datasets – even though they might show different conditions or domains), or performing imputation (e.g., by means of interpolation, zero padding) to have the same support for each data point \cite{GraphEmbed10,GraphEmbed11,GraphEmbed12,GraphEmbed13,GraphEmbed14,GraphEmbed15}. 
This is to guarantee that the manifold resulting from the embedding operation would be smooth and differentiable, so that the tangent plane to each $d$-dimensional data point on the manifold surface would resemble the Euclidean space in $\mathbb{R}^d$ \cite{ReprLearn1,ReprLearn5,ReprLearn4,ReprLearn2,GraphEmbed7,GraphEmbedTheory1,GraphEmbedTheory2,GraphEmbedTheory3}.  

Hence, 
using state-of-the-art embedding algorithms on datasets with missing data 
would lead to a dramatic reduction the semantic capacity of the analysis (e.g., accepting that we reach a lower level of detail in the characterisation of the given dataset with respect to the richness of detail that the dataset could actually offer), or even information loss (if the imputation process is imposing artefacts in the data to be then projected onto the manifold structure) \cite{GraphEmbed10,GraphEmbed11,GraphEmbed12,GraphEmbed13,GraphEmbed14,GraphEmbed15,ImpaintingRisks,GEOMDL_limits_2,GEOMDL_limits_3,GEOMDL_limits_4,GEOMDL_limits_7}. 

To overcome these major issues, 
graph embedding algorithms have to show the following properties: 
\begin{itemize}
	\item \textbf{P1}: adapt to the diverse characteristics of the data (multiple resolutions, different acquisition conditions, etc.), dealing missing and corrupted data without jeopardizing the accuracy and reliability of the analysis; 
	\item \textbf{P2}: provide an analytical and theoretical description of the information processing, so to ensure a thorough interpretation of the analysis and consistency of the data analysis methods.
\end{itemize}

\begin{figure}[htb]
	\centering
	\includegraphics[width=1\columnwidth]{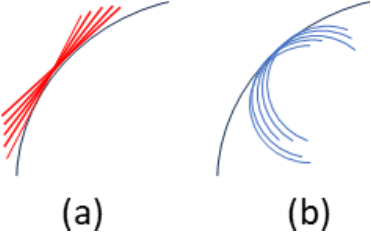} 
	\caption{Visual representation of the paradigm change provided by this work. In soft manifolds, the tangent planes to the points in the manifold are not forced to be Euclidean flat as in state-of-the-art manifolds (case (a)). On the contrary, they can be modeled accordingly to the characteristics of each data point as realizations of the fluid diffusion mechanism, hence showing non-linear shapes (case (b)) and higher flexibility to track the properties of the given dataset also in case of missing data.}
	\label{fig_nonFlat_planes}
\end{figure}

Therefore, 
it is crucial to identify a novel manifold structure that would support the definition of a continuous space in case of datasets with missing values. 
To this aim, we recall how changing the paradigm of information propagation modelling - from heat diffusion to fluid diffusion - is able to address the needs of data analysis (especially when graph-based) when the considered datasets are showing the characteristics \textbf{C1}-\textbf{C3}
\cite{fluid_ICASSP23,fluid_arxiv}. 
This implies removing the assumption that tangent planes to the point living on soft manifolds would be Euclidean flat 
(such as those in Figure \ref{fig_nonFlat_planes}(a)). 
On the contrary, they can be modeled accordingly to the characteristics of each data point as realizations of the fluid diffusion mechanism, hence showing non-linear shapes (see Figure \ref{fig_nonFlat_planes}(b)) and higher flexibility to track the properties of the given dataset also in case of missing data.

Following this logic, in this work we consider that the target manifold to be used for graph embedding would be defined as a result of a dynamic process based on the fluid diffusion mechanism. 
This choice allows us to encode the structural and topological properties of the graphs in 
the intrinsic fabric of manifold. 
In fact, the distance between data points would depend on a number of factors 
(i.e., relevance of each feature in the computation of the edge weights; rate by which the diffusion can take place; confidence associated with the diffusion across sections of the dataset) that are characteristics of the considered graph. 
According to the analogy proposed in \cite{fluid_arxiv}, 
we can assume that a manifold could be described by a number of \textit{structural} properties (e.g., curvature), but also of \textit{mechanical} characteristics \cite{Poincare}. 
Hence, we can assume that a manifold could be characterized by a \textit{shape}, and also by the \textit{materials} it consists of.
This would mean that the graph characteristics could be mapped on corresponding mechanical features (i.e., conductivity, diffusion rate, diffusivity) of the material that the target manifold would consist of at local and global level \cite{Poincare}. 

This allows us to propose a first discrimination between manifolds. 
In particular: 
\begin{itemize}
	\item if the distance between two data points is linearly dependent on the difference that can be computed feature-wise over all the records associated with the considered data points, then the conductivity, diffusivity, and diffusion rate of the manifold does not change. Thus, the material of the manifold would not change across the data points, and we
	can say that the manifold used to represent the data is a \textbf{hard} manifold;
	\item when instead 
	the number of features changes across the dataset (as in case of missing data), the conductivity, diffusivity, and diffusion rate, would change across the dataset, hence the material of the induced manifold. In this case, we can say that the considered manifold is a \textbf{soft} manifold.
\end{itemize} 
 
To achieve this goal, we propose to investigate the characteristics of the tangent spaces of soft manifolds, and their corresponding mapping onto the soft manifold topology. 
This requires to explore the nonlinearity of the fluid diffusion system, and how it affects the definition of the manifold, in particular when graph embedding is the ultimate task to be achieved. 
With this in mind, the next Section reports the main steps that we have taken to define the optimization algorithm that can be used to embed graphs with missing data on soft manifolds.


\section{Methods}
\label{sec_meth_all}

This Section introduces the main aspects of the proposed approach for graph embedding based on soft manifolds. 
First, 
we describe the principles of soft manifolds, and how they can be used for graph embedding. 
Then, we introduce the graph embedding algorithm based on soft manifolds.

\subsection{Principles of soft manifolds}
\label{sec_meth_soft}

The inherent ability of the fluid diffusion mechanism to model the interactions of data points when complex datasets are considered (in particular for graph representation \cite{fluid_arxiv,fluid_ICASSP23}) sets fluid dynamics model at the forefront to identify a continuous space that can be used to characterize the details of datasets showing missing data, and ultimately be used to extract information from them for diverse purposes and tasks. 
For reader's sake,  
we summarize in Appendix \ref{app_meth_fluid} the main elements of fluid dynamics model that can be expressed in terms of the Fokker-Planck equation, that are further detailed in \cite{fluid_arxiv}.

The difficulty in retrieving a robust manifold representation that can incorporate the characteristics of fluid dynamics model and address the problem of missing data 
lies in the discrepancy between the different supports associated with each data point, which result in the  incompatibility
between the space in which the spectral calculation has to be performed and the
space in which solutions depend differentiably on their initial data \cite{InvariantManifolds}. 
This difficulty is further emphasized when aiming at the goal of obtaining a solid scheme for graph embedding that would show the properties \textbf{P1}-\textbf{P2} as in Section \ref{sec_back}. 

To achieve a thorough understanding of the differentiable dependencies across data points, it is important to establish how the manifolds describing the continuous space that can embed the dataset under exam could address the presence of solutions with variable supports whilst guaranteeing the possibility to perform differential analysis on them. 
These spaces can be described as realizations of dynamical systems. In particular, we focus our attention to the fluid diffusion mechanism in porous media, which can be described by the following 
Fokker-Planck equation 
\cite{fluid_FokkerPlanck1,fluid_FokkerPlanck2}: 


\begin{equation}
	\frac{\partial p(\textbf{x},t)}{\partial t} = -  \nabla \cdot \left \{ \textbf{v}(\textbf{x}, {\cal K}) p(\textbf{x},t) \right \} 
	+ \nabla \cdot \tilde{\textbf{B}}(\textbf{x})\nabla p(\textbf{x},t),
	\label{eq_FokkerPlanck_fluid}
\end{equation}

\noindent where $\textbf{x} \in \mathbb{R}^n$ identifies one of the $N$ samples characterized by $n$ features in a given dataset, hence modelling one realization of the fluid diffusion mechanism in the $n$-dimensional space; $p(\textbf{x},t)$ is the transition probability for the fluid diffusion system (ultimately used to calculate the distance between data points in the resulting continuous space); 
$\textbf{v}(\textbf{x}, {\cal K})$ is a length-$n$ vector, 
models the transport velocity of information between data points. 
Taking inspiration from operational scenarios \cite{FluidFlow_CompFluidDyn_1,FluidFlow_CompFluidDyn_2,FluidFlow_CompFluidDyn_3,FluidFlow_CompFluidDyn_4}, 
the velocity between discrete sample \textit{i} and sample \textit{j} is computed as $v_{ij} = - || {\cal K}_{ij:}^T \odot (\textbf{x}_i - \textbf{x}_j) ||_2 \in [0,1]$, where ${\cal K}_{ij:}$ is the length-$n$ row vector collecting the third dimension elements of the conductivity tensor $\cal K$ on the $(i,j)$ coordinates and $\odot$ is the Hadamard product; 
the $n \times n$ matrix $\tilde{\textbf{B}}(\textbf{x})$ models rate of a diffusion process starting  from a given $\textbf{x}$ along the $n$ dimensions of the space; 
and $\nabla = [\frac{\partial}{\partial x_i}]_{i=1, \ldots, n}$. 
For more details, the reader is encouraged to refer to Appendix \ref{app_meth_fluid}. 

In general, the domains on which
solutions of systems such as (\ref{eq_FokkerPlanck_fluid}) live differ from the one that are obtained when considering the higher order asymptotic solutions \cite{InvariantManifold2,InvariantManifolds}. 
As such, the intrinsic stationarity of the higher order asymptotic solution of (\ref{eq_FokkerPlanck_fluid}) (also known as Barenblatt solution) cannot be directly used to work on self-similarity of variables, hence leading to closed form analytical expression of the actual interplay among data points in the continuous space \cite{InvariantManifold2,InvariantManifolds}. 

To overcome this issue, a nonlinear change of dependent and independent variables (thus leading to reparametrization of the solution subspace) was recommended \cite{InvariantManifold2}. 
In this way, the problem (\ref{eq_FokkerPlanck_fluid}) - which by default would not have fixed boundary conditions - would become a partial
differential equation on a fixed and finite domain \cite{InvariantManifolds}. 

To fully appreciate the effect of the proposed change of variables, it is worth rewriting (\ref{eq_FokkerPlanck_fluid}) as a function of a new variable $s(\textbf{x},t)$ such that $s(\textbf{x},t) = p(\textbf{x},t) ||\textbf{v}(\textbf{x},{\cal K})||_2$. 
Thus, the diffusion mechanism in (\ref{eq_FokkerPlanck_fluid}) would result
as follows:

\begin{equation}
	\frac{\partial }{\partial t} \frac{s(\textbf{x},t)}{||\textbf{v}(\textbf{x},{\cal K})||_2} = \Delta(\varphi(s(\textbf{x},t))), 
	\label{eq_FokkerPlanck_fluid_porousmedium}
\end{equation}

\noindent where $\Delta(y) = \nabla \cdot \nabla(y)$ is the Laplace operator, and $\varphi(s(\textbf{x},t)) = \int \tilde{\textbf{B}}(\textbf{x})\nabla s(\textbf{x},t)||\textbf{v}(\textbf{x},{\cal K})||_2^{-1} - \nabla \frac{\textbf{v}(\textbf{x},{\cal K})}{||\textbf{v}(x,{\cal K})||_2}s(\textbf{x},t) d\textbf{x}$. It is worth noting that the velocity term is always greater than zero when diffusion is possible, hence avoiding any issue with division by 0 for any factor in (\ref{eq_FokkerPlanck_fluid_porousmedium}).  

At this point, we can perform an initial change of variables on the spatial coordinates from $\textbf{x}$ to $\textbf{u}$, where \cite{InvariantManifold2,InvariantManifolds}:

\begin{equation}
	\textbf{u} = \frac{\textbf{x}}{\sqrt{2\frac{s(\textbf{x},t)}{||\textbf{v}(\textbf{x},{\cal K})||_2} + ||\textbf{x}||_2}}. 
	\label{eq_softman_changevar}
\end{equation}

With this change of variables, the solutions of (\ref{eq_FokkerPlanck_fluid_porousmedium}) (hence of (\ref{eq_FokkerPlanck_fluid})) are now confined onto the unit ball, and their positions depend also on the mechanical characteristics of the manifold, here identified by the conductivity and diffusivity of each point in every direction of the $n$-dimensional space according to the definitions in (\ref{eq_flowrate})  \cite{InvariantManifold2,InvariantManifolds}. 
Also, the space of the asymptotic solutions of (\ref{eq_FokkerPlanck_fluid_porousmedium}) is actually an half sphere in the (\textbf{u},t) space \cite{InvariantManifolds}. 
With this in mind, we define $y(\textbf{u},t)$ as the distance between the point (\textbf{x},$\sqrt{\frac{s(\textbf{x},t)}{||\textbf{v}(\textbf{x},{\cal K})||_2}}$) and the half sphere, i.e.:

\begin{equation}
	1 + y(\textbf{u},t) = \sqrt{2\frac{s(\textbf{x},t)}{||\textbf{v}(\textbf{x},{\cal K})||_2} + ||\textbf{x}||_2}.
	\label{eq_softman_changevar2}
\end{equation}

With this additional change of variables, the asymptotic solutions to (\ref{eq_FokkerPlanck_fluid_porousmedium}) are transformed onto the constant function set to 1 in all dimensions. 
Consequently, the perturbations to these solutions are reduced to 0 \cite{InvariantManifolds}. 
Furthermore, the graph of $y$ is diffeomorphic to the graph of $s$, and consequently of $p$ \cite{InvariantManifold2,InvariantManifolds}. 

It is also convenient to consider an additional definition of the variables in (\ref{eq_softman_changevar}) and (\ref{eq_softman_changevar2}) as follows:

\begin{equation}
	\frac{s(\textbf{x},t)}{||\textbf{v}(\textbf{x},{\cal K})||_2} - r(\textbf{x}) = y(\textbf{u},t) + \frac{1}{2} y(\textbf{u},t)^2,
	\label{eq_softman_changevar3}
\end{equation}

\noindent where $r(\cdot) = \frac{1}{2}(1- ||\cdot||^2_2)$. 
Thus, the equation in (\ref{eq_FokkerPlanck_fluid_porousmedium}) can be written as for the $y = y(\textbf{u},t)$ variable as follows:

\begin{equation}
	\frac{\partial y}{\partial t} - \nabla_\textbf{u} \cdot (r \nabla_\textbf{u} y) = \psi(\textbf{v},|\nabla_\textbf{u} y|^2),
	\label{eq_FokkerPlanck_fluid_porousmedium2}
\end{equation}

\noindent where $\nabla_\textbf{u} = [\frac{\partial}{\partial u_i}]_{i=1, \ldots, n}$, and $\psi$ is a positive kernel depending on $|\nabla_\textbf{u} y|^2$ and on the operator ${\cal L}_\textbf{u} = -\Delta + \textbf{u} \cdot \nabla$ \cite{InvariantManifolds}. 

At this point, it is possible to construct smooth solutions to (\ref{eq_FokkerPlanck_fluid_porousmedium2}) for initial data in $C^{0,1}$. 
Also, the solutions are prevented to degenerate by the Lipschitz norm of the initial datum, making them uniformly controlled in any $C^k$ norm \cite{InvariantManifolds}. 
Finally, it is possible to prove \cite{InvariantManifold2,InvariantManifolds} that solutions depend analytically on the initial conditions, and that the problem in (\ref{eq_FokkerPlanck_fluid_porousmedium2}) is \textit{well posed} \cite{InvariantManifolds}. 
Specifically, let us consider two constants $ \tau>0$ and $\upsilon>0$, and let us define ${\cal B}_{\tau,\upsilon} = \{ y \in C^{0,1}: ||y||_{L^\infty} \leq \tau, ||y||_{\Lip} \leq \upsilon \}$. 
Then, we can state that there is a unique solution $y$ to (\ref{eq_FokkerPlanck_fluid_porousmedium2}) for every $y^* \in {\cal B}_{\tau,\upsilon}$. 
Indeed, this solution is smooth and depends analytically on $y^*$. 
Moreover, $||y||_{L^\infty} + ||y||_{\Lip} <1$, and $t^{\alpha +|\beta|} | \frac{\partial^\alpha}{\partial t} \frac{\partial^\beta}{\partial \textbf{u}} \nabla_\textbf{u} y(\textbf{u},t)| \lesssim ||y^*||_{\Lip}$ for $\alpha \in \mathbb{N}_0$, $\beta \in \mathbb{N}_0^n$, and $\forall (\textbf{u},t) \in B_1(0) \times ]0,+\infty[$, where $B_1(0)$ is the unit ball centred at the origin of the $n$-dimensional space \cite{InvariantManifold2,InvariantManifolds}. 

This result allows us to draw some interesting conclusions \cite{InvariantManifolds}:
\begin{itemize}
	\item  the solutions of (\ref{eq_FokkerPlanck_fluid_porousmedium2}) are stable, with a finite rate of convergence;
	\item the spatial translations of $y$ can be described in terms of first order corrections to the asymptotic solutions of (\ref{eq_FokkerPlanck_fluid_porousmedium2});
	\item if the center of mass of the fluid dynamics system is fixed at the origin, spatial translations of $y$ become negligible and the dynamics is governed by affine transformations;
	\item studying the spectrum of the eigenvalues of the dynamic system in (\ref{eq_FokkerPlanck_fluid_porousmedium2}), it is possible to assess how the fluid dynamic system is robust to dilations, i.e., invariant to time shifts.  
\end{itemize}

This set-up allows us to define a class of manifolds that are defined as realizations of the fluid dynamical system in (\ref{eq_FokkerPlanck_fluid_porousmedium2}), particularly by considering the asymptotic behavior of the solutions of (\ref{eq_FokkerPlanck_fluid_porousmedium2}). 
It is indeed possible to define invariant and differentiable continuous maps on these manifolds: the details for this proof are reported in Appendix \ref{app_map}.  

In the next paragraphs, we will introduce the major properties of these manifolds, especially focusing on 
the manifold metric, and the definition of a distance between to points on the manifold.

\subsubsection{Metric}

Especially, the definition of the metric is instrumental for the characterization of the decay of the solutions and their boundaries \cite{InvariantManifold2,InvariantManifolds}. 
Considering the discussion that we summarized in the initial part of this Section, taking a look to the derivation we proposed in (\ref{eq_FokkerPlanck_fluid_porousmedium}), one could be drawn to think that the manifold induced by the fluid dynamic systems in (\ref{eq_FokkerPlanck_fluid}) could be described in terms of classic Riemannian terms. 
As such, we could try to make use of the tools provided in the technical literature of theoretical studies of Riemannian manifolds to derive a metric that could resemble the outcome of a system based on heat diffusion \cite{InvariantManifold2}. 

However, taking a closer look to (\ref{eq_FokkerPlanck_fluid_porousmedium2}) and the definition of the different terms on both sides of the equation, we can observe that a metric based on the classic heat kernel would fail to track the deep characteristics of the manifold resulting from the fluid dynamic system. 
In particular, using a metric based on the heat kernel would fail to describe the intrinsic nonlinearity (of second order) of the system in (\ref{eq_FokkerPlanck_fluid_porousmedium2}) at the boundary of the domain \cite{InvariantManifolds}. 
To achieve the goal of full and solid characterization of the properties of the manifold induced by (\ref{eq_FokkerPlanck_fluid_porousmedium2}), we would need indeed to use a Carnot - Caratheodory distance defined on the unit ball (thanks to the change of variables we proposed in (\ref{eq_softman_changevar})-(\ref{eq_softman_changevar3})) \cite{InvariantManifold2}. 

Specifically, to characterize the nonlinearity at the boundaries of the domain, we could consider to \textit{weigh} each infinitesimal portion of the manifold. 
In other terms, let us consider a positive function $\Upsilon$ on the manifold ${\cal S}$ induced by (\ref{eq_FokkerPlanck_fluid_porousmedium2}).  
Also, let us consider $d\Vol_\textbf{q}$ the volume element of the manifold ${\cal S}$ when a metric $\textbf{q}$ is set. 
Moreover, we can define $d \gamma = \Upsilon d\Vol_\textbf{q}$. 
Then, the Laplacian operator on the local coordinates of  the $({\cal S}, \textbf{q}, \gamma)$ system can be written as follows: 

\begin{equation}
	\Delta_\gamma y = \frac{1}{\Upsilon \sqrt{\det \textbf{q}}} \sum_{i,j=1}^{n} \partial_i (\breve{q}_{ij} \Upsilon \sqrt{\det \textbf{q}} \partial_j y),
	\label{eq_soft_manifold_metric}
\end{equation}

\noindent where $\breve{\textbf{q}} = \{\breve{q}_{ij}\}_{(i,j) \in \{1, \ldots, n\}^2}$ is the inverse of $\textbf{q}$, and $\Delta_\gamma(\cdot) = \nabla_\gamma \cdot \nabla_\gamma(\cdot)$, with $\nabla_\gamma = [\frac{\partial}{\partial \gamma_i}]_{i=1, \ldots, n}$. 

With this in mind, let us consider $\textbf{q}$ to be the conformally flat Riemannian metric on the Euclidean unit ball $B_1(0)$, i.e., $\textbf{q} = r^{-1}(d\textbf{x})^2$, and let us set $\Upsilon = r^{n/2}$. 
Under these conditions, (\ref{eq_FokkerPlanck_fluid_porousmedium2}) can be written as a special form of the heat diffusion equation as follows:

\begin{equation}
	\frac{\partial y}{\partial t} - \Delta_\gamma y = \psi.
	\label{eq_soft_manifold_metric2}
\end{equation}
 
At this point, let us define $\nu = \arcsin(||\textbf{x}||)$. 
Thus, we can write as follows: 

\begin{equation}
	\frac{1}{2} \textbf{q} = \frac{1}{2} r^{-1}(d\textbf{x})^2 = (d \nu)^2 + (\tan \nu)^2 (d\ell_{\textbf{S}^{n-1}})^2.
	\label{eq_soft_manifold_metric3}
\end{equation}

\noindent where $d\ell_{\textbf{S}^{n-1}}$ is the length element on the $(n-1)$-dimensional unit sphere. 

Hence, we can say that the metric $\textbf{q}$ to be associated with the manifold ${\cal S}$ resulting from the dynamic system in (\ref{eq_FokkerPlanck_fluid_porousmedium2}) is intuitively half-way the Euclidean metric $(d\textbf{x})^2$ (used in the Riemannian manifold analysis) and the metric on the hyperbolic Poincare' disk $r^{-2}(d\textbf{x})^2$ \cite{InvariantManifolds,InvariantManifold2}. 

With this in mind, we can define the distance between two points in the $({\cal S}, \textbf{q})$ system. 
We report its derivation in the following Section. 

\subsubsection{Distance between two points}

Using conformally flat coordinates is instrumental to derive the expression of the distance between two points on the manifold ${\cal S}$. 
Thus, we consider the metric $\textbf{q} = r^{-1}(d\textbf{x})^2$ on the unit ball $B_1(0)$. 
In this scenario, the distance 
defined on $B_1(0)$ can be considered an intrinsic distance for diffusion as in (\ref{eq_FokkerPlanck_fluid_porousmedium2}), i.e., the square distance provides an estimate of the time scale at which the fluid is exchanged between two points \cite{InvariantManifolds}. 

With this in mind, and defining $\Lambda(\Gamma)= \int_{\zeta_1}^{\zeta_2} \frac{|\Gamma'(\xi)|}{\sqrt{1-|\Gamma(\xi)|^2}} d \xi$, where $\Gamma(\zeta_1) = \textbf{u}_1$ and $\Gamma(\zeta_2) = \textbf{u}_2$,  we can derive the intrinsic distance of (\ref{eq_FokkerPlanck_fluid_porousmedium2}) (hence, the distance between two points on the manifold ${\cal S}$) as follows:

\begin{equation}
	d_{{\cal S}}(\textbf{u}_1, \textbf{u}_2) = \inf \{ \Lambda(\Gamma) : \Gamma \: \mathrm{joins} \: \textbf{u}_1 \: \mathrm{and} \: \textbf{u}_2\},
\label{eq_soft_manifold_distance}	
\end{equation}

Minimizing $\Lambda(\Gamma)$ is indeed equivalent to minimize $E(\Gamma)$, i.e.:

\begin{equation}
	E(\Gamma) = \int_{\zeta_1}^{\zeta_2} \frac{|\Gamma'(\xi)|^2}{1-|\Gamma(\xi)|^2} d \xi. 
	\label{eq_soft_manifold_distance2}
\end{equation}

\noindent This is because $\Lambda(\Gamma)^2 \leq (\zeta_2 - \zeta_1)E(\Gamma)$ (by Holder inequality \cite{InvariantManifold2}). 
Also, $\Lambda(\Gamma)^2 = (\zeta_2 - \zeta_1)E(\Gamma)$ if $\Gamma$ is a geodesic curve, i.e., $\frac{|\Gamma'(\xi)|}{\sqrt{1-|\Gamma(\xi)|^2}}$ is constant \cite{InvariantManifolds}. 

The functional in (\ref{eq_soft_manifold_distance2}) is convex, so we can find a unique minimum. 
It is hence convenient to parametrize the minimizer by the arclength, so we can write $|\Gamma'|^2 = 1 - |\Gamma|^2$ and $\left( \frac{ \Gamma'}{1-|\Gamma|^2} \right)' = \frac{ \Gamma}{1-|\Gamma|^2}$ according to the Euler-Lagrange equations.  

The geodesic curves over the manifold ${\cal S}$ are defined over the 2-dimensional plane induced by $\textbf{u}_1$, $\textbf{u}_2$, and the origin of the $n$-dimensional space, although this plane is not anymore flat, e.g.,  for the classic Riemannian case \cite{InvariantManifolds,HeterManifold}. 
In general, $\Gamma$ can be expressed as $\Gamma = [w_1, w_2, 0, \ldots, 0] \in \mathbb{R}^n$. 
As a result, we can write as follows: 
\begin{equation}
	|\Gamma'|^2 =  1 - |\Gamma|^2  =  (w_1')^2 + (w_2')^2 = 1 - w_1^2 - w_2^2   
	\label{eq_soft_manifold_distance3}
\end{equation}

Also, the $\left( \frac{ \Gamma'}{1-|\Gamma|^2} \right)'$ term can be decomposed in these factors:

\begin{eqnarray}
	\left( \frac{ w_1'}{1-w_1^2-w_2^2} \right)' & = &  \frac{ w_1}{1-w_1^2-w_2^2}  \nonumber \\
	\left( \frac{ w_2'}{1-w_1^2-w_2^2} \right)' & = &  \frac{ w_2}{1-w_1^2-w_2^2}  
	\label{eq_soft_manifold_distance4}
\end{eqnarray}

At this point, it is important to notice that these equations can be solved by \textit{hypocycloid curves}, i.e., traces of fixed points on small
circles of radius $\rho \leq 1/2$ that roll along the interior boundary of the unit ball. 
In other terms, considering that we are working on the unit ball $B_1(0)$, we can write:

\begin{eqnarray}
w_1(t) & = &  (1-\rho) \cos \left( t\sqrt{\frac{\rho}{1-\rho}} \right)  + \rho \cos \left( t\sqrt{\frac{1-\rho}{\rho}} \right) \nonumber \\
w_2(t) & = &  (1-\rho) \sin \left( t\sqrt{\frac{\rho}{1-\rho}} \right)  - \rho \sin \left( t\sqrt{\frac{1-\rho}{\rho}} \right) 
	\label{eq_soft_manifold_distance5}
\end{eqnarray}
 
Solving the problem (\ref{eq_soft_manifold_distance})-(\ref{eq_soft_manifold_distance2}) is not trivial: indeed, it can be cumbersome to analytically derive the expression of the intrinsic distance between two points of ${\cal S}$. 
However, 
it is possible to prove that the geodesic distance can be  approximated by a semimetric, so that 
the following holds \cite{InvariantManifolds}:

\begin{equation}
	d_{{\cal S}}(\textbf{u}_1,\textbf{u}_2) \sim \mathring{d}_{{\cal S}}(\textbf{u}_1,\textbf{u}_2) =  \frac{||\textbf{u}_1 - \textbf{u}_2||}{\sqrt{||\textbf{u}_1-\textbf{u}_2||} + \sqrt{r(\textbf{u}_1)} + \sqrt{r(\textbf{u}_2)}}. 
	\label{eq_soft_manifold_distance_equiv}
\end{equation}

Therefore, in this work we will consider to use the expression in (\ref{eq_soft_manifold_distance_equiv}) as the function to use to compute the distance between two points on the manifold ${\cal S}$.

\subsubsection{Definition of soft manifold}

Taking into account the properties we previously discussed, we can state that a manifold $({\cal S}, \textbf{q})$ is a $n$-dimensional \textit{soft} manifold if it can be considered as a result of a $n$-dimensional dynamical system as in (\ref{eq_FokkerPlanck_fluid_porousmedium2}), and that is equipped with a metric $\textbf{q} = r^{-1}(d\textbf{x})^2$. 
In this case, 
the manifold ${\cal S}$ can be arranged on a unit ball $B_1(0)$ in the $n$-dimensional space that summarizes the structural and mechanical characteristics of the fluid diffusion model in (\ref{eq_FokkerPlanck_fluid_porousmedium2}), but 
the tangent hyperplane to any point of ${\cal S}$ is not flat (as e.g., in the case of Riemannian manifolds). 
As such, the distances between points must be computed along hypocycloid curves, hence resulting in the geodesic of the form as in (\ref{eq_soft_manifold_distance_equiv}). 

We can therefore define \textit{hard} manifolds those for which the mechanical characteristics of the datasets (e.g., conductivity, diffusivity in (\ref{eq_FokkerPlanck_fluid})) do not change across the considered data points. 

\subsection{Graph embedding based on soft manifolds}

\begin{algorithm}
	\caption{Graph embedding on soft manifold}\label{euclid}
	\begin{algorithmic}[1]
		\State Given a dataset $\textbf{X} = \{\textbf{x}_i\}_{i=1, \ldots, N}$
		\State \textit{Perform} change of variables as in (\ref{eq_softman_changevar})-(\ref{eq_softman_changevar2})
		\For{$(i,j) \in \{1,\ldots, N\}^2$}
		\State \textit{compute} $d_{\cal G}^2(\textbf{x}_i,\textbf{x}_j)$ as in (\ref{eq_fluid_trans_prob_pij})
		\State \textit{compute} $d_{\cal S}(\textbf{u}_i, \textbf{u}_j)$ as in  (\ref{eq_soft_manifold_distance_equiv})
		\EndFor
		\State \textbf{end for}
		\State \textit{Define} $\mathsf{L}_d(\textbf{U})$ as in (\ref{eq_graph_embed_3})
		\For{$i \in \{1, \ldots, N\}$}
		\State \textit{compute} $\bar{A}^i$ as in (\ref{eq_graph_embed_7})
		\State compute $\bar{{\cal A}}^i$ as in (\ref{eq_graph_embed_11})
		\EndFor 
		\State \textbf{end for}
		\State \textit{Define} $\mathsf{L}_g (\textbf{U})$ as in (\ref{eq_graph_embed_12})
		\State minimize $\mathsf{L}(\textbf{U}) = \mathsf{L}_d(\textbf{U}) + \kappa \mathsf{L}_g(\textbf{U})$ 
	\end{algorithmic}
\end{algorithm}

In this Section, we report the main steps of the algorithm for graph embedding based on the soft manifold that have been previously introduced. 
Specifically, given a graph ${\cal G} = ({\cal V}, {\cal E})$ consisting of $N$ nodes $\textbf{X} = \{x_i\}_{i=1,\ldots,N}$ (included in the ${\cal V}$) and a set of edges ${\cal E}$ connecting them, we aim to find an embedding $\forall \textbf{x} \in {\cal V}$ that can be written as follows:

\begin{equation}
	f:\textbf{x} \mapsto \textbf{u} \in {\cal S},
	\label{eq_graph_embed_1}
\end{equation}

\noindent where ${\cal S}$ is a soft manifold with the characteristics that have been defined in Section \ref{sec_meth_soft}. 

Taking advantage of the spherical symmetry intrinsic to the soft manifolds, and inspired by the approach to graph embedding proposed in \cite{HeterManifold}, we build the graph embedding algorithm as a minimization operation based on a composite loss function $\mathsf{L}(\textbf{U})$, where $\textbf{U} = \{\textbf{U}_i\}_{i=1, \ldots, N}$ identifies the nodes of $\textbf{X}$ embedded on ${\cal S}$. 
Indeed, the loss function consists of two factors, i.e., 

\begin{equation}
	\mathsf{L}(\textbf{U}) = \mathsf{L}_d(\textbf{U}) + \kappa \mathsf{L}_g(\textbf{U}),
	\label{eq_graph_embed_2}
\end{equation} 

\noindent where $\kappa$ is a scale parameter for regularization. 
In this work, we chose to minimize the loss function $\mathsf{L}$ by means of the stochastic gradient descent (SGD) algorithm \cite{HeterManifold}. 

The two factors $\mathsf{L}_d$ and $\mathsf{L}_g$ aim to grasp different characteristics of the embedding, so that the resulting projection on the soft manifolds can reliably incorporate the major properties of the considered graph. 
These terms are explicitly defined in the following Sections. 

\subsubsection{Definition of $\mathsf{L}_d$}

The factor $\mathsf{L}_d$ aims to model the long range interactions across nodes, hence accounting for the average distance distortion \cite{HeterManifold}. 
As such, $\mathsf{L}_d$ can be written as follows:

\begin{equation}
	\mathsf{L}_d(\textbf{U}) = \sum_{i,j} \left| \frac{d_{\cal S}^2(\textbf{u}_i, \textbf{u}_j)}{d_{\cal G}^2(\textbf{x}_i,\textbf{x}_j) + \epsilon_d} - 1\right|,
	\label{eq_graph_embed_3} 
\end{equation}

\noindent where $d_{\cal S}(\textbf{u}_i, \textbf{u}_j)$ is defined as in (\ref{eq_soft_manifold_distance_equiv}), whereas the distance between nodes $i$ and $j$ in graph ${\cal G}$ $d_{\cal G}^2(\textbf{x}_i,\textbf{x}_j)$ is set to (\ref{eq_fluid_trans_prob_pij}). $\epsilon_d$ is a constant to avoid numerical instabilities.

\subsubsection{Definition of $\mathsf{L}_g$}

The $\mathsf{L}_g$ term aims to assess how close the local geometry of the manifold ${\cal S}$
around the embedded nodes resembles that of the graph ${\cal G}$ \cite{HeterManifold}. 
On the fluid graph derived in \cite{fluid_arxiv,fluid_ICASSP23}, and according to the definition of the Forman curvature \cite{HeterManifold}, 
a metric to model the local geometry of the neighborhood of node $i$ ${\cal N}(i)$ by summing the area of the trinagles radially induced by the elements of ${\cal N}(i)$ regularly located around node $i$. 
Specifically, given node $i$ and ${\cal N}(i) = \{i_1, i_2, \ldots, i_{|{\cal N}(i)|} \}$, assuming $d_{{\cal G}}(i,i_1) \leq d_{{\cal G}}(i,i_2), \leq \ldots \leq d_{{\cal G}}(i,i_{|{\cal N}(i)|})$, let us consider to uniformly place these nodes radially around node $i$ at a distance proportional to each corresponding $d_{{\cal G}}$ over directions separated by an angle $\theta_i = 2\pi/|{\cal N}(i)|$. 
Then, we can count the area of the triangle induced by nodes $i - i_{j} - i_{\tilde{j}}$, $j \in \{1, \ldots, |{\cal N}(i)|\}$ and $\tilde{j} =  j\mod |{\cal N}(i)|+1$, which can be defined as follows:

\begin{equation}
	A^i_{j,\tilde{j}} = \frac{1}{2} d_{{\cal G}}(i,i_j) d_{{\cal G}}(i,i_{\tilde{j}}) \sin \theta_i.
	\label{eq_graph_embed_4}
\end{equation}

Thus, we can define the area that is spanned by the whole local neighborhood of $i$ as follows:

 \begin{equation}
 	A^i = \sum_{j=1}^{|{\cal N}(i)|} A^i_{j,\tilde{j}}
 	\label{eq_graph_embed_5}
 \end{equation}

\begin{figure}[htb]
	\centering
	\includegraphics[width=1\columnwidth]{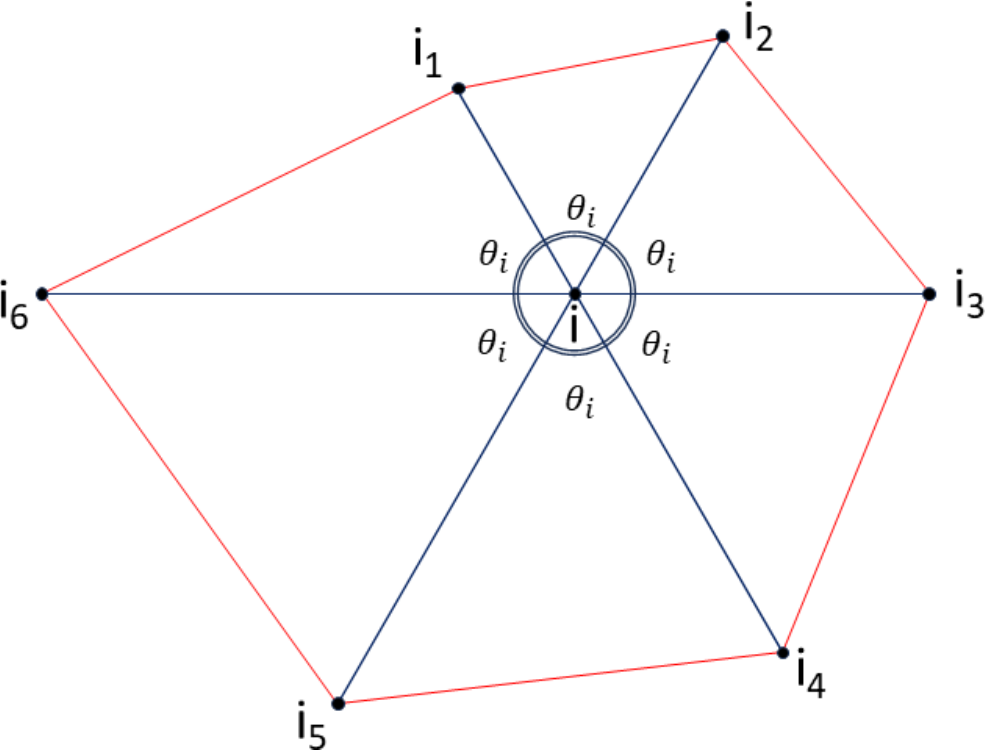} 
	\caption{Visual representation of $A^i$ as defined in (\ref{eq_graph_embed_5}) when $|{\cal N}(i)|=6$ (hence $\theta_i = \pi/3$): the value of $A^i$ is obtained by calculating the area of the region within the red line. }
	\label{fig_Lg_Ai}
\end{figure}

The visual representation of $A^i$ when $|{\cal N}(i)|=6$ is displayed as the area of the region contained within the red lines in Figure \ref{fig_Lg_Ai}. 
It is worth noting that if $|{\cal N}(i)|=1$ then $A^i=0$.
Also, if $d_{\cal G}(i,i_j) \rightarrow 0$ $\forall j$, then $A^i \rightarrow 0$. 
Therefore, we can state that $A^i$ would tend to 0 for nodes that are outliers or showing a high degree of centrality to the graph. 
Finally, the maximum area that can be spanned in the given graph ${\cal G}$ is as follows:

\begin{equation}
	A^* =  \frac{{|{\cal N}(i)|}}{2} d_{\cal G}^{*^2} \sin \theta_i,
	\label{eq_graph_embed_6}
\end{equation}

\noindent where $d_{\cal G}^{*} = \max_{i,j \in \cal V} d_{\cal G}(i,j)$. 

With this in mind, we can set the metric that we can use to qunatify the characteristics of the local geometry of the neighborhood of node $i$ in the graph ${\cal G}$ as follows:

\begin{equation}
	\bar{A}^i = A^i/A^*. 
	\label{eq_graph_embed_7}
\end{equation}

Let us now consider the manifold ${\cal S}$. 
Specifically, let us assume that the point $i$ and one of the points in its neighborhood $i_j$ induce an angle $\varphi^i_j$ between the radii that connect these two points (placed onto the unit ball) to the origin of the $n$-dimensional space, i.e., $\varphi^i_j = d_{\cal S}(i,i_j)/R\stackrel{R=1}{=} d_{\cal S}(i,i_j)$. 
On the azimut polar coordinate, the $j$-th point of ${\cal N}(i)$ is radially located at an angle $\vartheta^i_j$, and separated to the point $(j+1)$ by an angle $\vartheta_i = 2\pi/|{\cal N}(i)|$. 
Thus, the area of the spherical sector induced by the  by nodes $i - i_{j} - i_{\tilde{j}}$, $j \in \{1, \ldots, |{\cal N}(i)|\}$, $\tilde{j} =  j\mod |{\cal N}(i)|+1$, with respect to the origin of the unit ball can be written as follows:

\begin{eqnarray}
	{\cal A}^i_{j,\tilde{j}} = & \int_{\vartheta^i_j}^{\vartheta^i_{\tilde{j}}} \int_{\varphi^i_j}^{\varphi^i_{\tilde{j}}} \rho^2 \sin \phi \: d\phi \: d\theta \nonumber \\
	= & \rho^2 (\vartheta^i_{\tilde{j}} - \vartheta^i_j) (\cos \varphi^i_j  - \cos \varphi^i_{\tilde{j}}) \\
	 = & \vartheta_i (\cos \varphi^i_j  - \cos \varphi^i_{\tilde{j}}) \nonumber
	\label{eq_graph_embed_8}
\end{eqnarray}

Therefore, the area of the spherical sector spanned by the whole neighborhood ${\cal N}(i)$ is as follows:

 \begin{equation}
	{\cal A}^i = \sum_{j=1}^{|{\cal N}(i)|} {\cal A}^i_{j,\tilde{j}}
	\label{eq_graph_embed_9}
\end{equation}

It is worth noting that the maximum area of  spherical sector over the unit ball as induced by the manifold ${\cal S}$ is as follows:

\begin{eqnarray}
	{\cal A}^* = & \int_0^{2\pi} \int_0^{\varphi^*} \rho^2 \sin \phi \: d\phi \: d\theta \nonumber \\
	 = & 2\pi (1 - \cos \varphi^*),
	 \label{eq_graph_embed_10}
\end{eqnarray}

\noindent where $\varphi^*$ is the inclination angle in polar coordinates that is associated with the maximum distance $d_{{\cal S}}^*$ that can be spanned by a pair of points of the manifold ${\cal S}$. 
With this in mind, we can then compute the normalized area of the spherical sector associated with the neighborhood of point $i$ of the manifold ${\cal S}$ as follows:

\begin{equation}
	\bar{{\cal A}}^i = {\cal A}^i/{\cal A}^*. 
	\label{eq_graph_embed_11}
\end{equation}

We can now determine the formula associated with the loss term $\mathsf{L}_g$ as follows:

\begin{equation}
	\mathsf{L}_g (\textbf{U})= \sum_{i} \left| \frac{\bar{{\cal A}}^i}{\bar{A}^i + \epsilon_g} - 1\right|,
	\label{eq_graph_embed_12}
\end{equation}

\noindent where $\epsilon_g$ is a constant to avoid numerical instabilities.

\subsection{Node prediction}
\label{sec_meth_nodepred}

Once the manifold is derived as a result of the graph embedding procedure, node prediction can be conducted. 
The labelling can be performed by identifying the best matching mapping of the nodes to be predicted over the shape of the produced manifold \cite{MoNet}. 
As such, for the node prediction task we used a GCN network equipped with a soft-max layer, so to produce a probability distribution of each node to be mapped on the manifold structure. 
At this point, 
learning is done by
minimizing the standard logistic regression cost \cite{MoNet}. 

%

\section{Experimental results and discussion}
\label{sec_result}

We conducted several experiments to show how the proposed approach for graph embedding based on soft manifolds is beneficial to address the limits of modern data analysis and in particular the case of missing data. 
To this aim, we tested the scheme we introduced in Section \ref{sec_meth_all} on datasets from diverse research fields.  
These datasets show high diversity in terms of heterogeneity of sensing platforms and acquisition strategies (e.g., multimodal and unimodal datasets; static and dynamic records). 
Thus, the considered datasets display various characteristics in terms of noise distributions across the samples, as well as of semantics of the data (i.e., inter- and intra-class relationships). 

This high degree of diversity make these datasets a great fit to analyze the robustness of the proposed  approach for graph embedding in case of missing data. 
We tested this point by simulating the occurrence of missing data (by setting subsets of features to a null value).
We assessed the outcomes we obtained by comparing with state-of-the-art methods, i.e., the methods reported in \cite{HeterManifold} ('Heterogeneous Manifold') and \cite{GraphEmbed14} ('Symmetric space'). 

In this Section, we first summarize the main characteristics of the datasets we have taken into account. 
Then, 
we report the performance of the proposed graph embedding algorithm, and its effects on node prediction task. 

\subsection{Datasets}
\label{sec_exp_res}
We tested the proposed approach on four very diverse datasets, focusing on two different research fields: product co-purchasing, and remote sensing.

\subsubsection{Product co-purchasing}
\label{sec_exp_OGBAmazon}

First, we considered a dataset representing an Amazon product co-purchasing network \cite{data_OGBAmazon}. 
The dataset consists of more than 2.5M nodes, characterized by 100 features. 
47 classes of products are drawn onto this dataset, so to be used for the node prediction task.

\subsubsection{Multimodal remote sensing (RS)}
\label{sec_exp_RS}
We considered a multimodal 
dataset consisting of LiDAR and hyperspectral records acquired over the University of Houston campus and the neighboring urban area,  distributed for the 2013 IEEE GRSS Data Fusion Contest~\cite{data_houston}. 
Specifically:
\begin{itemize}
\item the size of the dataset is $N=$1905$\times$349 pixels, with spatial resolution equal to 2.5m;
\item the final dataset consists of $n$=151 features. The hyperspectral dataset includes 144 spectral bands ranging from 0.38 to 1.05 $\mu{m}$, whilst the LiDAR records includes one band and 6 textural features;
\item the available ground truth labels consists of $K_C=15$ land use classes. 
\end{itemize}


\color{black}

\subsection{Results}
\label{sec_eval}

In order to provide a thorough investigation of the actual impact of the proposed approach, we conducted several experiments at different levels. 
Specifically, we tested the proposed architecture in terms of graph embedding capacity, and accuracy in node prediction. 

\begin{figure}[htb]
	\centering
	\includegraphics[width=1\columnwidth]{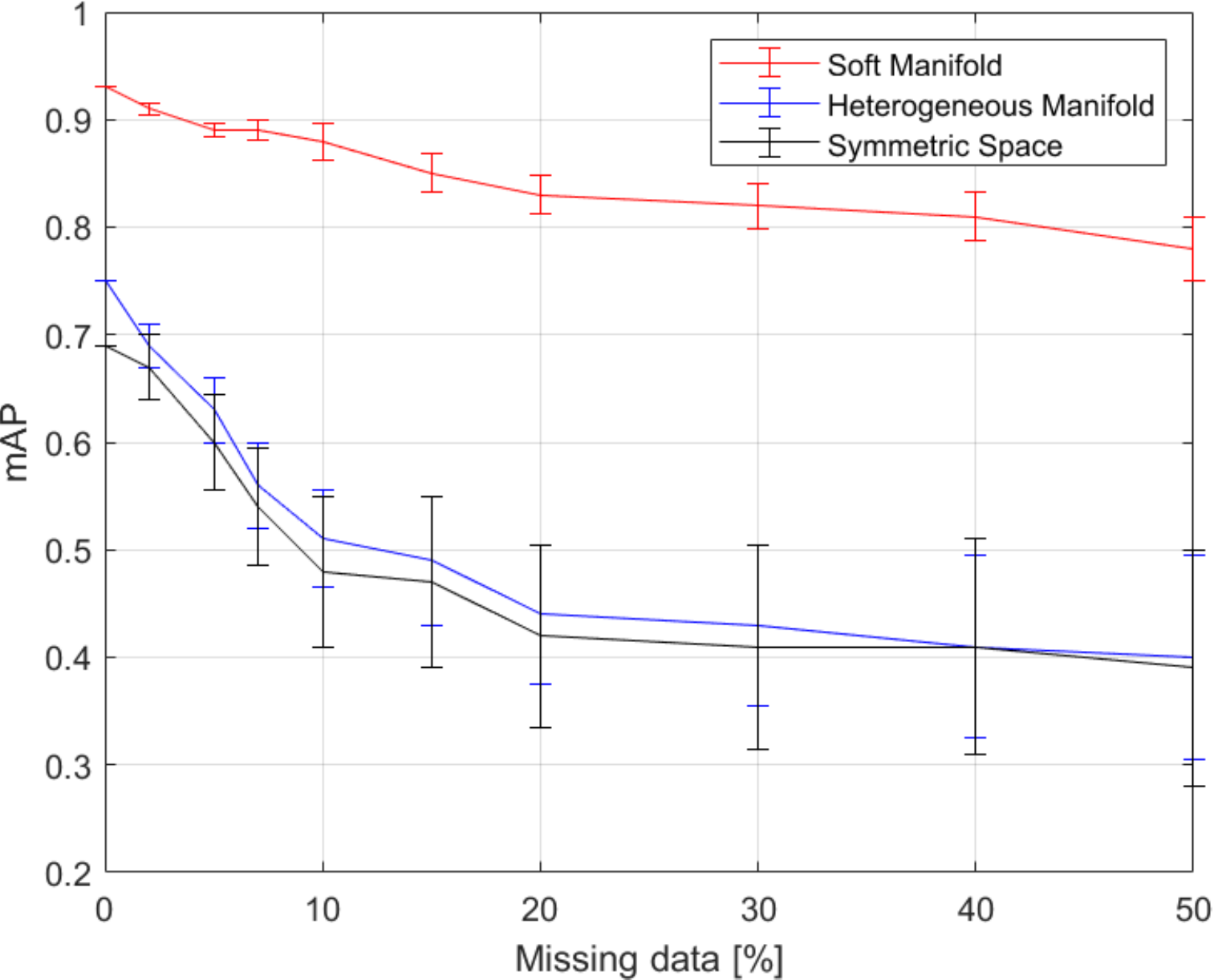} 
	\caption{Mean average precision (mAP) results obtained when performing graph embedding by means of the proposed algorithm ('Soft manifold', red line), the method proposed in \cite{HeterManifold} ('Heterogeneous manifold', blue line), and the method proposed in \cite{GraphEmbed14} ('Synthetic space', black line) over the product co-purchasing dataset in Section \ref{sec_exp_OGBAmazon} as a function of missing data. The error bar indicates the variance of the results achieved over 100 experiments for each missing data setting.}
	\label{fig_res_graphembed_OGBAmazon}
\end{figure}

\begin{figure}[htb]
	\centering
	\includegraphics[width=1\columnwidth]{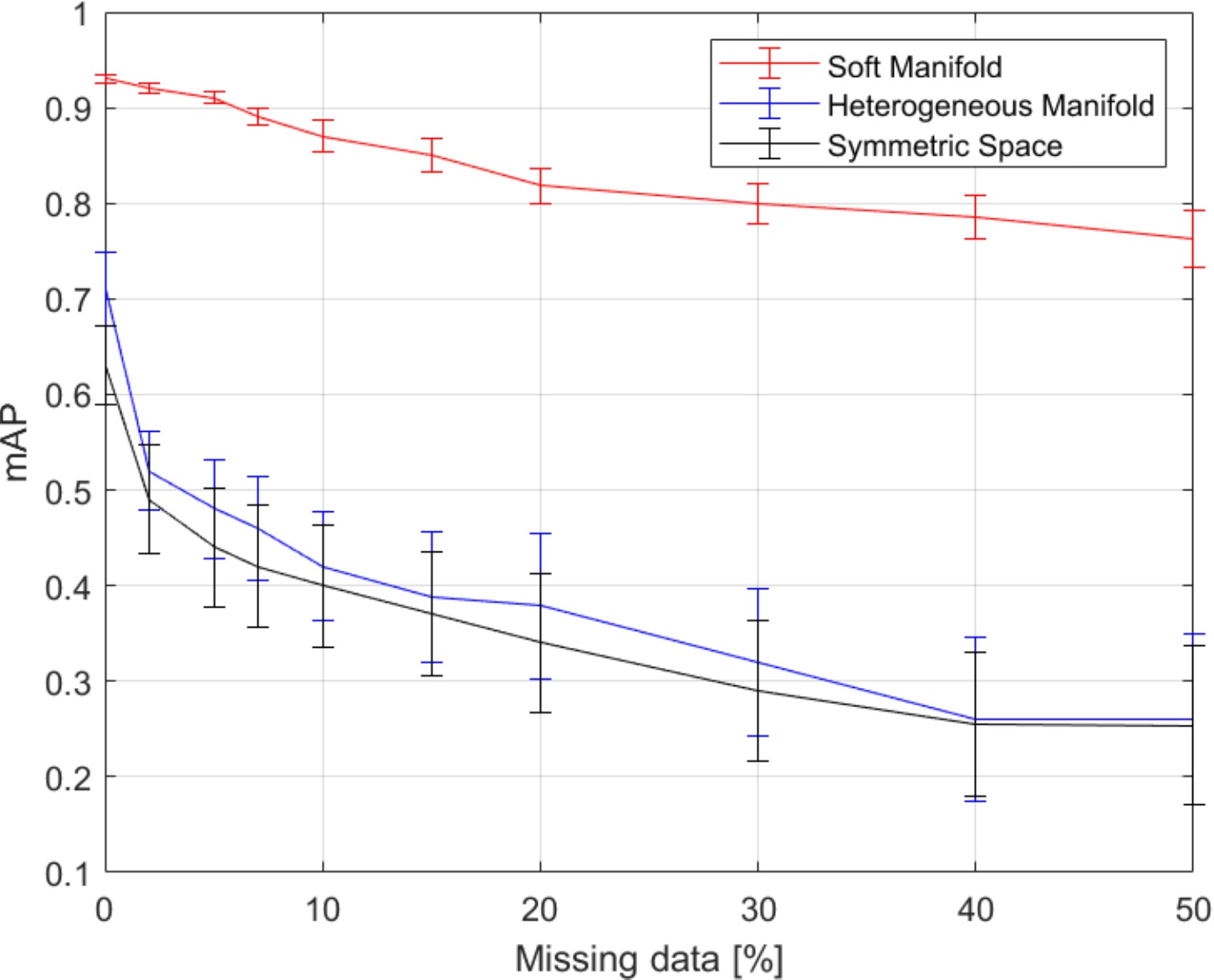} 
	\caption{Mean average precision (mAP) results obtained when performing graph embedding over the remote sensing  dataset in Section \ref{sec_exp_RS} as a function of missing data. The same notation of Figure \ref{fig_res_graphembed_OGBAmazon} applies here. }
	\label{fig_res_graphembed_RS}
\end{figure}

\begin{figure}[htb]
	\centering
	\includegraphics[width=1\columnwidth]{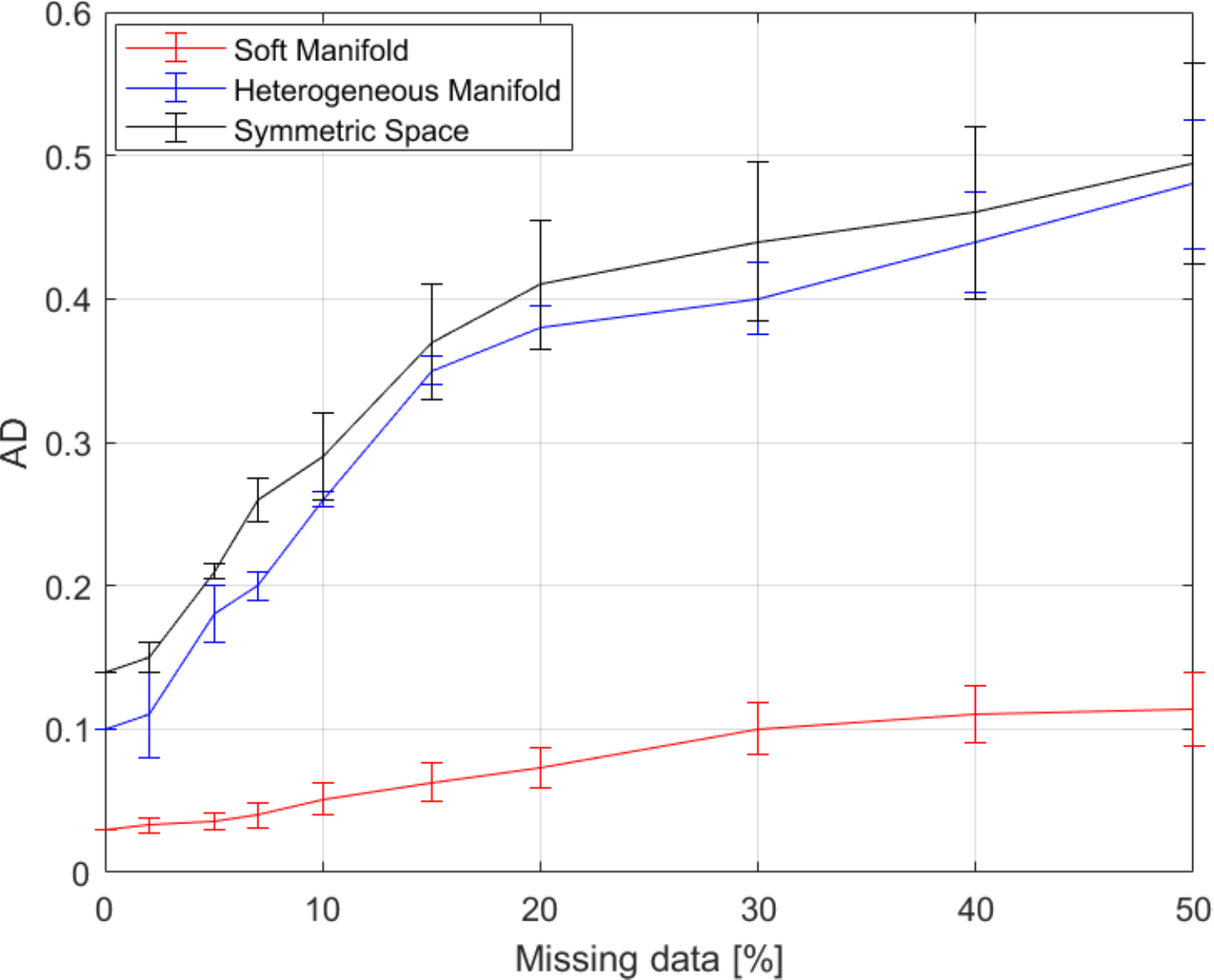} 
	\caption{Average distortion (AD) results obtained when performing graph embedding over the remote sensing  dataset in Section \ref{sec_exp_OGBAmazon} as a function of missing data. The same notation of Figure \ref{fig_res_graphembed_OGBAmazon} applies here. }
	\label{fig_res_graphembed_OGBAmazon_AD}
\end{figure}

\begin{figure}[htb]
	\centering
	\includegraphics[width=1\columnwidth]{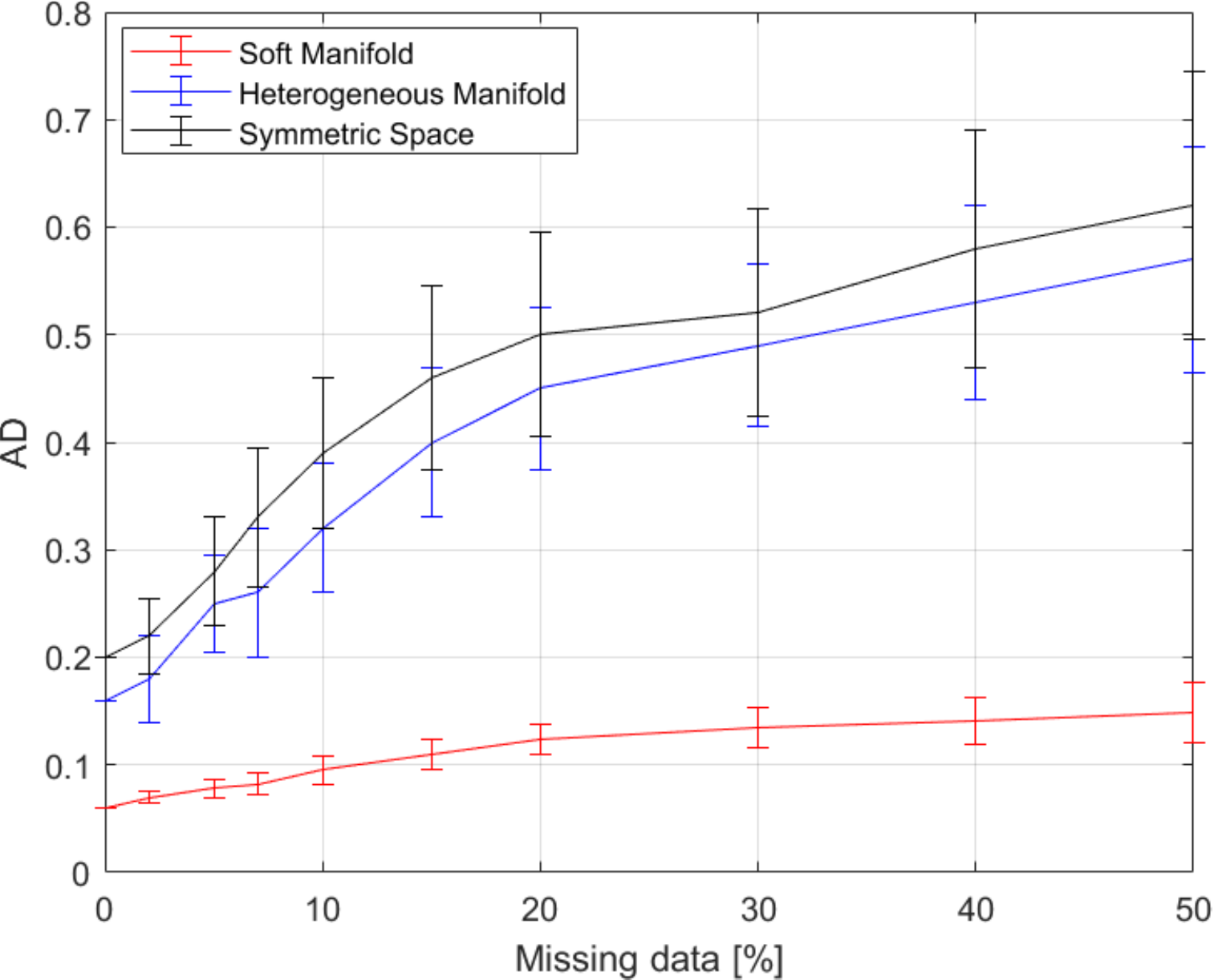} 
	\caption{Average distortion (AD) results obtained when performing graph embedding over the remote sensing  dataset in Section \ref{sec_exp_RS} as a function of missing data. The same notation of Figure \ref{fig_res_graphembed_OGBAmazon} applies here. }
	\label{fig_res_graphembed_RS_AD}
\end{figure}

\subsubsection{Graph embedding}

When considering the graph embedding task, we can explore the capacity of the proposed method by
evaluating two metrics:

\begin{itemize}
	\item the mean average precision of the graph embedding operation, which is defined as follows:
	\begin{equation}
		\mathsf{mAP}(f) = \frac{1}{N} \sum_{i \in {\cal V}} \frac{1}{d_i} \sum_{j \in {\cal N}(i)} \frac{|{\cal N}(i) \cap \mathfrak{R}_{ij}|}{|\mathfrak{R}_{ij}|}, 
	\end{equation}
	
	\noindent where $f$ is the embedding map, $d_i$ is the degree of node $i$, and $\mathfrak{R}_{ij}$ is the subset of nodes $z \in {\cal V}$ for which $d_{\cal M}(f(i),f(z)) \leq d_{\cal M}(f(i),f(j))$, being $d_{\cal M}$ is the distance on the manifold as defined according to each representation \cite{HeterManifold} (i.e., in this work, $d_{\cal M}$ is defined as in (\ref{eq_soft_manifold_distance_equiv})); 
	
	\item the average (distance) distortion of the graph embedding, which is defined as follows \cite{HeterManifold}:
	\begin{equation}
		\mathsf{AD}(f) = \frac{2}{N(N-1)} \sum_{i,j = 1}^N \left| 1 - \frac{d_{\cal M}(f(i),f(j))}{d_{\cal G}(i,j)}\right|.
	\end{equation}
\end{itemize}

Figures \ref{fig_res_graphembed_OGBAmazon} and \ref{fig_res_graphembed_RS} display the mAP results we achieved  when we considered the datasets in Section \ref{sec_exp_OGBAmazon} and \ref{sec_exp_RS}, respectively. 
Analogously, Figures \ref{fig_res_graphembed_OGBAmazon_AD} and \ref{fig_res_graphembed_RS_AD} report the AD results for these datasets. 
To obtain these results, we simulated the occurrence of a proportion missing features for each dataset. 
For each setup of missing data proportion, we conducted 100 experiments where the locations of the missing data in the considered datasets were randomly assigned. 

Taking a look to these results, it is possible to appreciate how the proposed approach outperforms the state-of-the-art strategies. 
In particular, it is worth noting that the improvement provided by the soft manifold-based graph embedding (both in mAP and AD terms) increases as the percentage of missing features available for each considered data point increases. 
This emphasizes that the proposed approach is actually able to tackle the lack of data availability when performing the manifold reconstruction task. 
It is also important to highlight that the variance of the method we introduced on the experiments we conducted over both datasets is significantly lower than those recorded for the methods in \cite{HeterManifold} and \cite{GraphEmbed14}. 
This shows that the soft manifold representation is robust to the different conditions identified by the experiments we conducted. 
As such, the proposed approach clearly outperforms the state-of-the-art methods for graph embedding on continuous spaces when the considered data points are showing missing features.
The benefit of our approach is further emphasized when performing node prediction, as displayed by the results we report in the following Section.

\subsubsection{Node prediction}

\begin{figure}[htb]
	\centering
	\includegraphics[width=1\columnwidth]{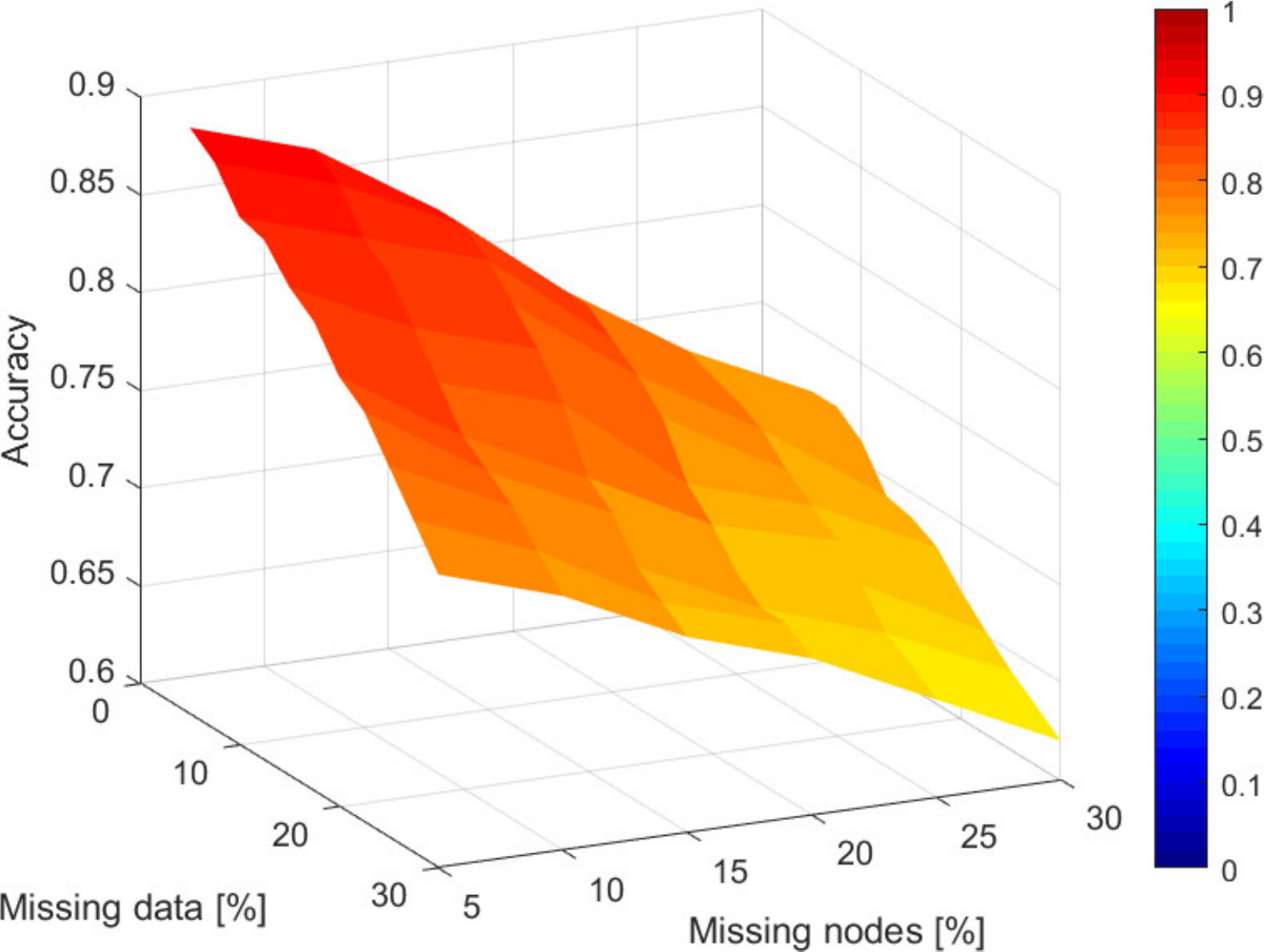} 
	\caption{Average accuracy results obtained when performing node prediction by means of the proposed architecture based on Soft Manifolds graph embedding over the product co-purchasing dataset in Section \ref{sec_exp_OGBAmazon} as a function of the percentage of missing nodes to be predicted and missing data. The accuracy results are displayed on a three dimensional surface according to the colormap on the right hand side.}
	\label{fig_res_nodepred_OGBAmazon_Soft3D}
\end{figure}

Node prediction is conducted according to the procedure described in Section \ref{sec_meth_nodepred}. 
Once again, to obtain these results, we simulated the occurrence of a proportion missing features for each dataset. 
For each setup of missing data proportion, we conducted 100 experiments where the locations of the missing data in the considered datasets were randomly assigned, as well as the nodes that needed to be predicted. 
As such, the results can be plotted in three-dimensional diagrams, such as Figure \ref{fig_res_nodepred_OGBAmazon_Soft3D}, where the accuracy achieved when analyzing the product co-purchasing dataset in terms of soft manifolds is displayed. 
To fully appreciate the characteristics of the results we obtained, in the following we break down these digrams in 2-D plots first, and then we analyze the gradient properties of these surfaces, so to provide a thorough description of the capacity and limits of the methods that were considered in this Section. 

\begin{figure}[htb]
	\centering
	\includegraphics[width=1\columnwidth]{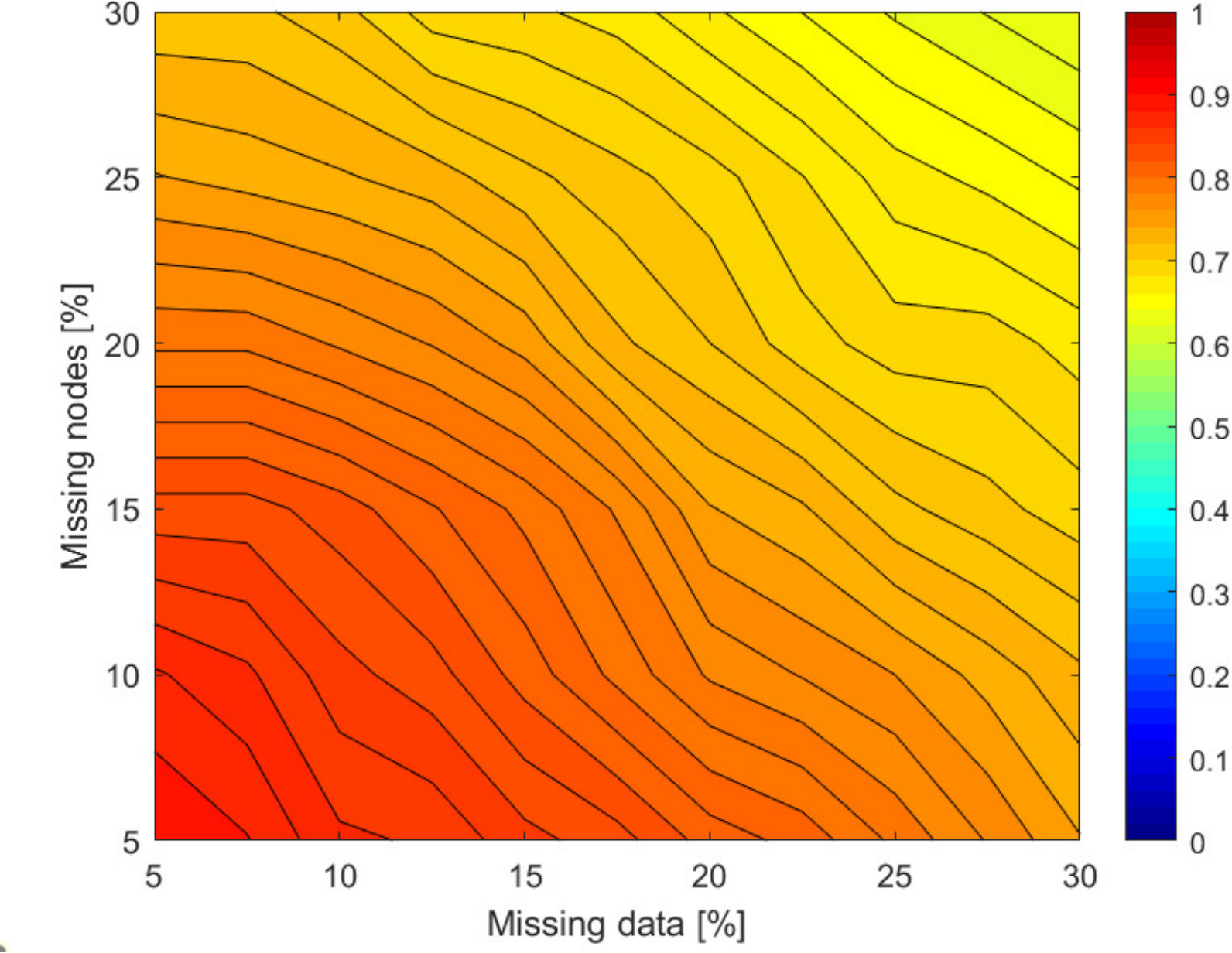} 
	\caption{Average accuracy results obtained when performing node prediction by means of the proposed architecture based on Soft Manifolds graph embedding over the product co-purchasing dataset in Section \ref{sec_exp_OGBAmazon} as a function of the percentage of missing nodes to be predicted and missing data. The accuracy results are displayed as contour curves with a color ranging in the [0,1] interval according to the colormap on the right hand side.}
	\label{fig_res_nodepred_OGBAmazon_Soft}
\end{figure}

\begin{figure}[htb]
	\centering
	\includegraphics[width=1\columnwidth]{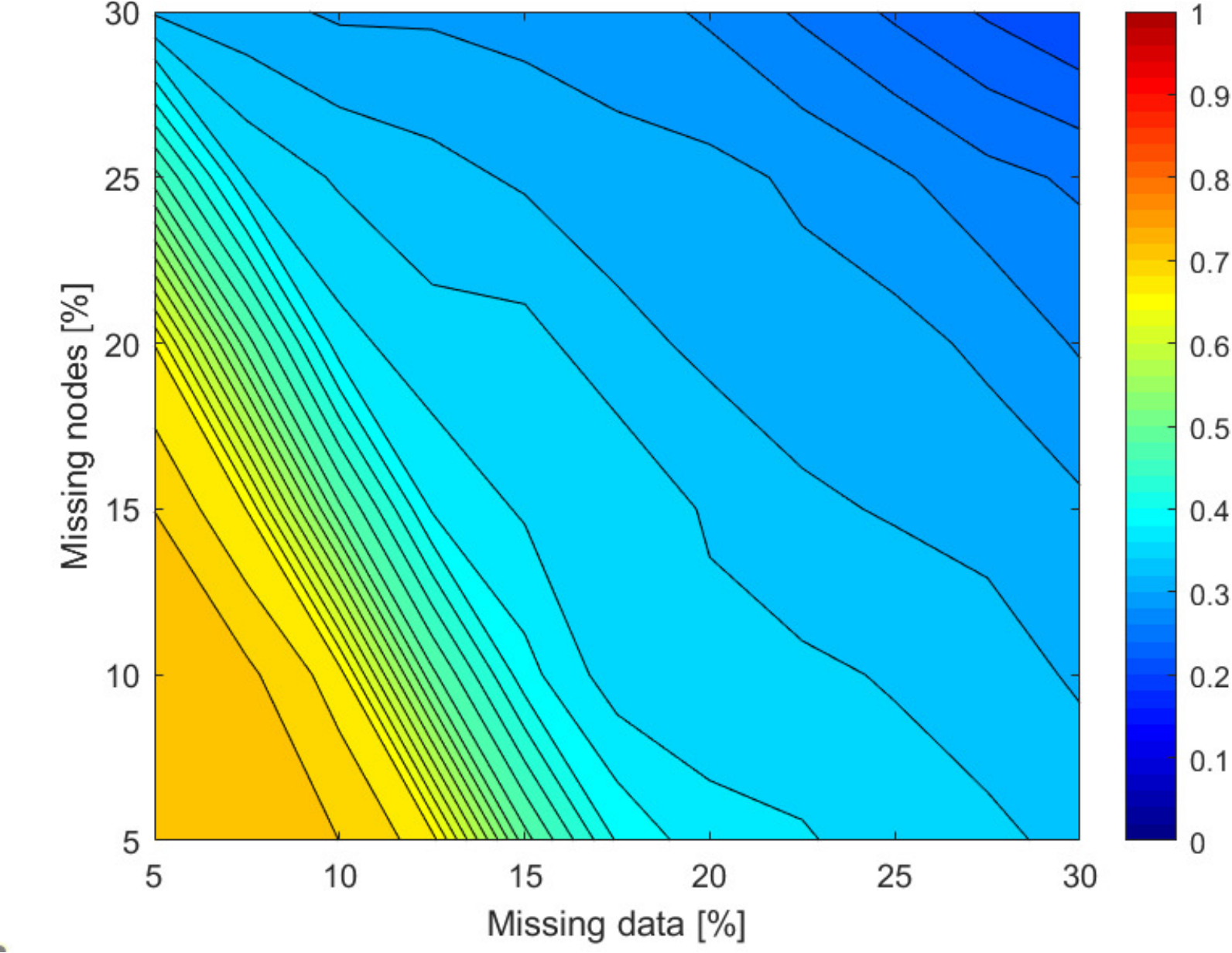} 
	\caption{Average accuracy results obtained when performing node prediction by means of the proposed architecture based on Heterogeneous Manifolds graph embedding \cite{HeterManifold} over the product co-purchasing dataset in Section \ref{sec_exp_OGBAmazon} as a function of the percentage of missing nodes to be predicted and missing data. The same notation of Figure \ref{fig_res_nodepred_OGBAmazon_Soft} applies here.}
	\label{fig_res_nodepred_OGBAmazon_Heter}
\end{figure}

\begin{figure}[htb]
	\centering
	\includegraphics[width=1\columnwidth]{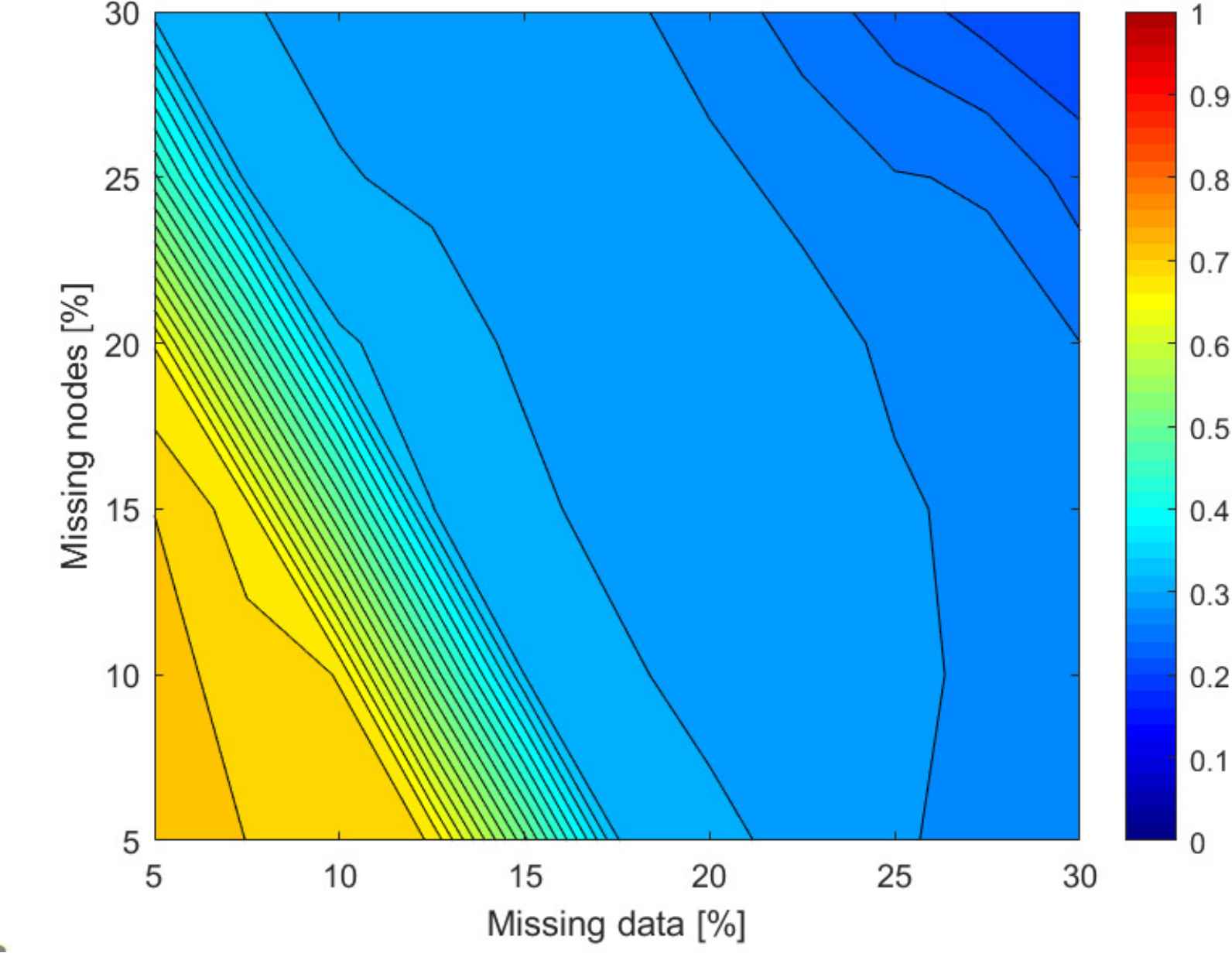} 
	\caption{Average accuracy results obtained when performing node prediction by means of the proposed architecture based on Symmetric Space graph embedding \cite{GraphEmbed14} over the product co-purchasing dataset in Section \ref{sec_exp_OGBAmazon} as a function of the percentage of missing nodes to be predicted and missing data. The same notation of Figure \ref{fig_res_nodepred_OGBAmazon_Soft} applies here. }
	\label{fig_res_nodepred_OGBAmazon_Symmetric}
\end{figure}

\begin{figure}[htb]
	\centering
	\includegraphics[width=1\columnwidth]{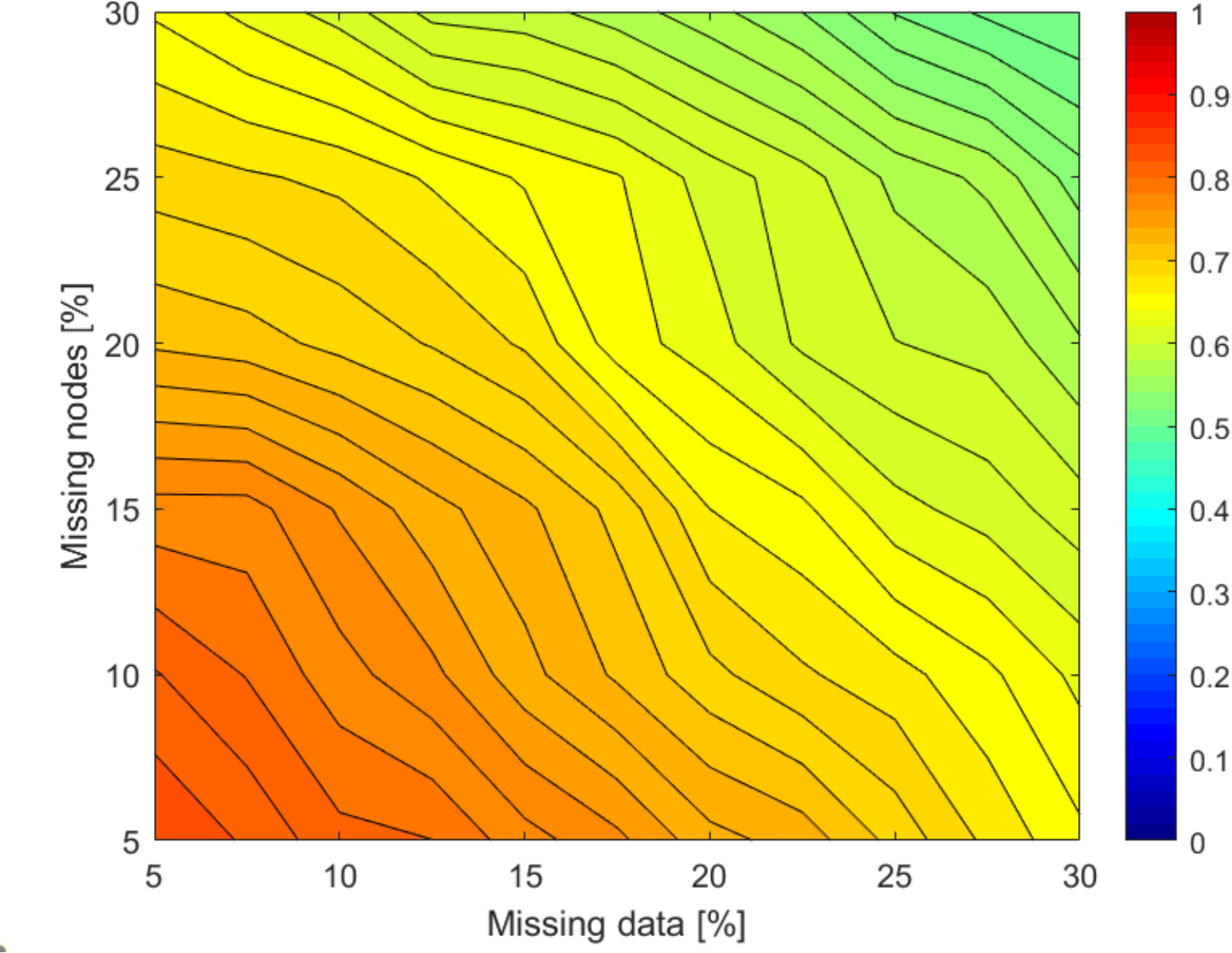} 
	\caption{Average accuracy results obtained when performing node prediction by means of the proposed architecture based on Soft Manifolds graph embedding over the remote sensing dataset in Section \ref{sec_exp_RS} as a function of the percentage of missing nodes to be predicted and missing data. The same notation of Figure \ref{fig_res_nodepred_OGBAmazon_Soft} applies here. }
	\label{fig_res_nodepred_RS_Soft}
\end{figure}

\begin{figure}[htb]
	\centering
	\includegraphics[width=1\columnwidth]{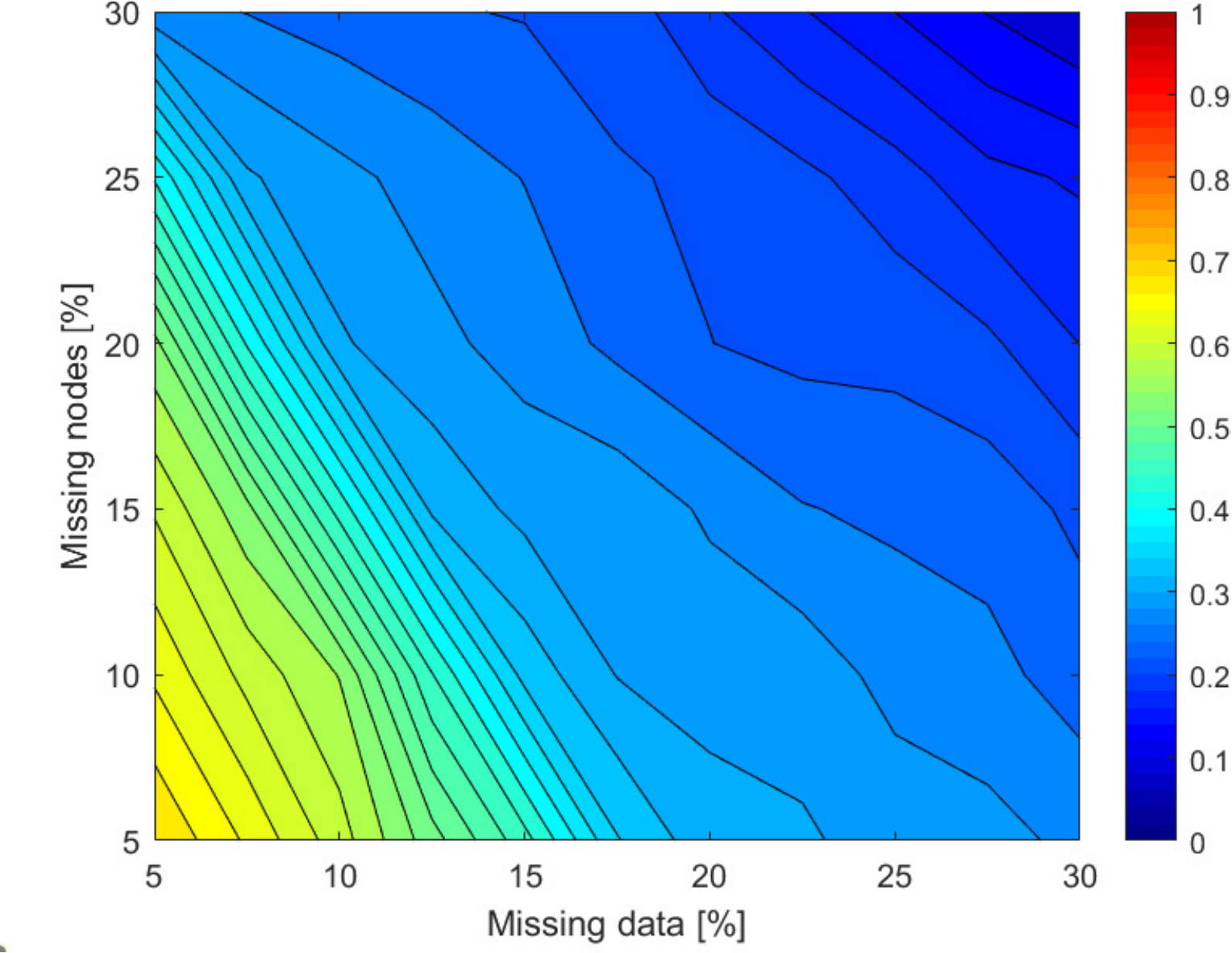} 
	\caption{Average accuracy results obtained when performing node prediction by means of the proposed architecture based on Heterogeneous Manifolds graph embedding \cite{HeterManifold} over the remote sensing dataset in Section \ref{sec_exp_RS} as a function of the percentage of missing nodes to be predicted and missing data. The same notation of Figure \ref{fig_res_nodepred_OGBAmazon_Soft} applies here. }
	\label{fig_res_nodepred_RS_Heter}
\end{figure}

\begin{figure}[htb]
	\centering
	\includegraphics[width=1\columnwidth]{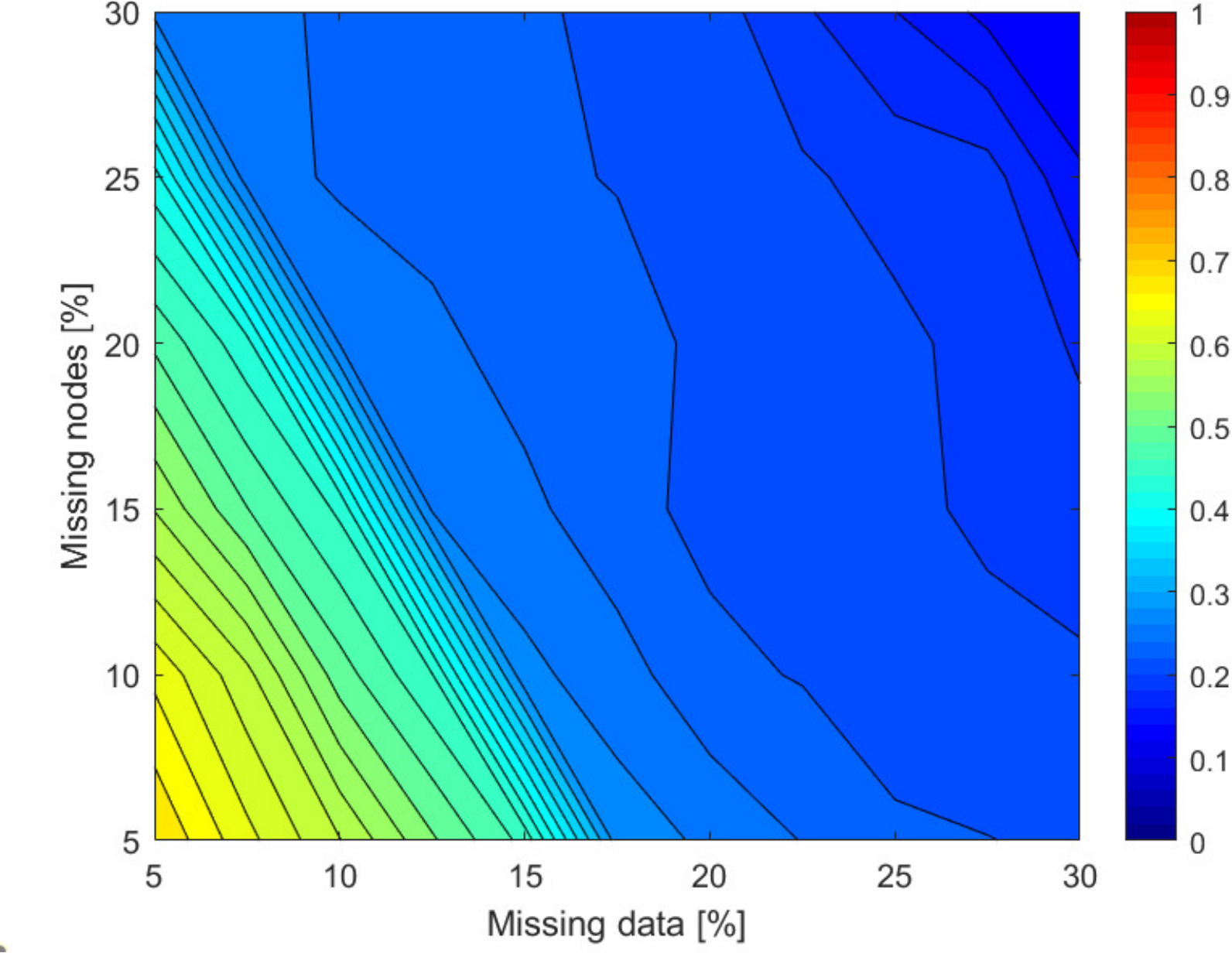} 
	\caption{Average accuracy results obtained when performing node prediction by means of the proposed architecture based on Symmetric Space graph embedding \cite{GraphEmbed14} over the product co-purchasing dataset in Section \ref{sec_exp_RS} as a function of the percentage of missing nodes to be predicted and missing data. The same notation of Figure \ref{fig_res_nodepred_OGBAmazon_Soft} applies here.  }
	\label{fig_res_nodepred_RS_Symmetric}
\end{figure}

Figures \ref{fig_res_nodepred_OGBAmazon_Soft}-\ref{fig_res_nodepred_OGBAmazon_Symmetric} display the results we achieved when considering the product co-purchasing dataset and the manifolds where embedded by using the algorithm proposed in this work, the method in \cite{HeterManifold} and the method in \cite{GraphEmbed14}, respectively. 
Analogously, the node prediction results we achieved by using the aforesaid methods to derive the manifold structure on the remote sensing dataset are reported in Figures \ref{fig_res_nodepred_RS_Soft}-\ref{fig_res_nodepred_RS_Symmetric}, respectively. 
The accuracy results are displayed as contour curves with a color ranging in the [0,1] interval according to the colormap on the right hand side of each Figure.

Let us take a look to these results. 
Once again, it is evident the strong impact of the proposed approach to derive the continuous space embedding the original graph on the extraction of reliable information from the given datasets. 
Specifically, it is possible to appreciate how using soft manifolds to represent the continuous space of the datasets under exam yields the best average accuracy for the node prediction task under all the conditions of missing nodes and missing data that have been considered. 

\begin{figure}[htb]
	\centering
	\includegraphics[width=1\columnwidth]{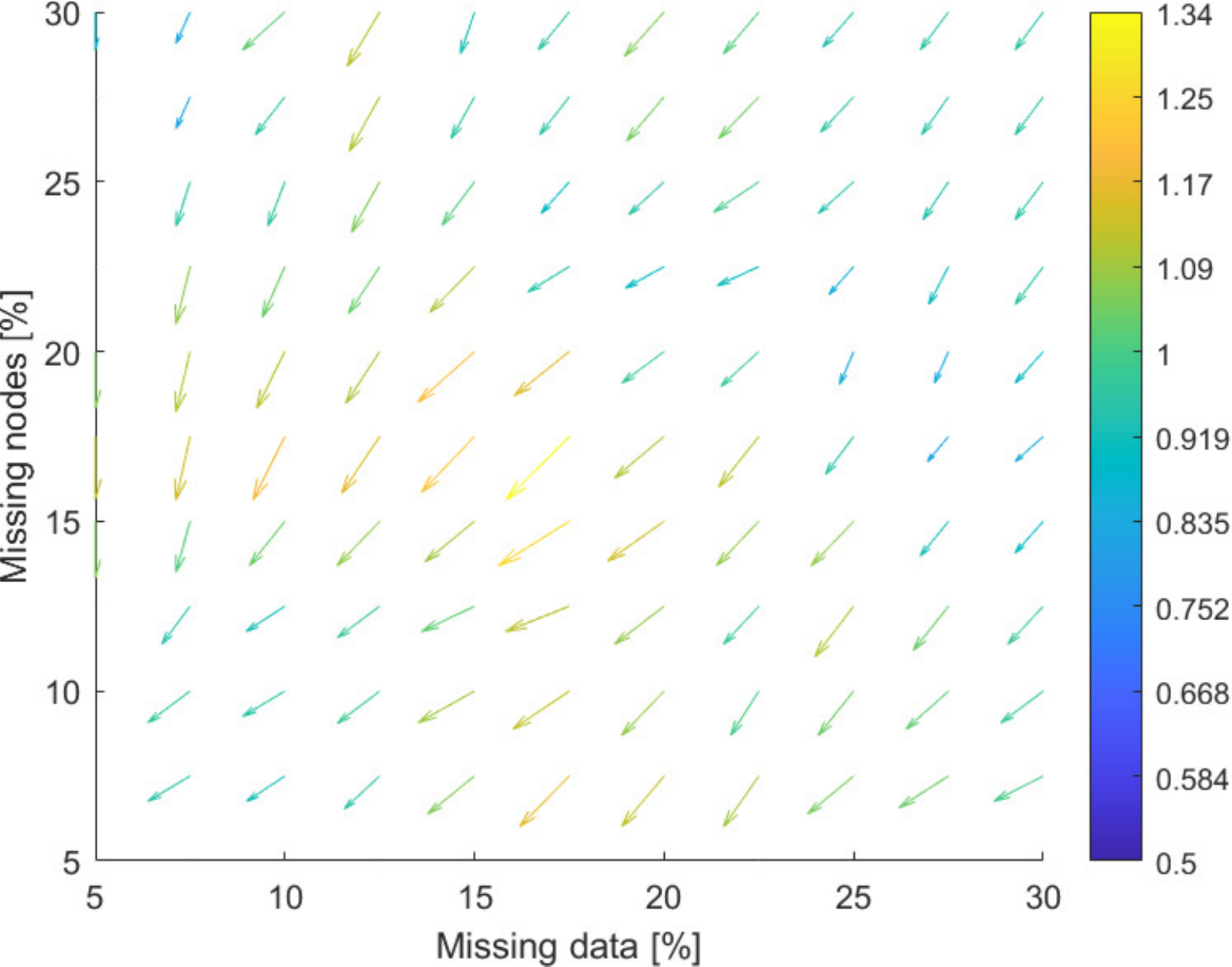} 
	\caption{Vector field representation of the ascending gradient for the average accuracy results reported in Fig. \ref{fig_res_nodepred_OGBAmazon_Soft}. The amplitude of the gradient is reported according to the color scale on the right hand side.}
	\label{fig_res_vecfield_OGBAmazon_Soft}
\end{figure}

\begin{figure}[htb]
	\centering
	\includegraphics[width=1\columnwidth]{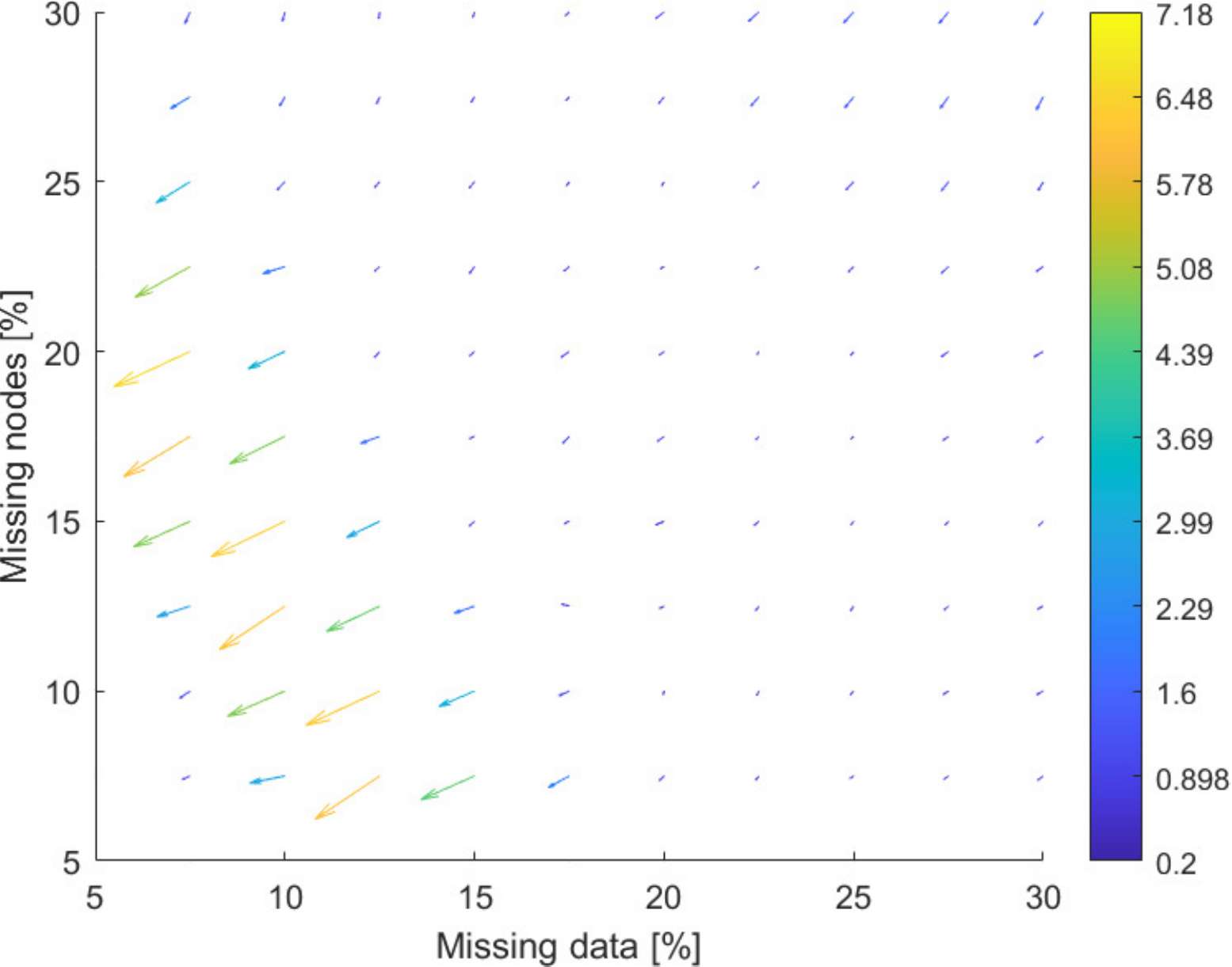} 
	\caption{Vector field representation of the ascending gradient for the average accuracy results reported in Fig. \ref{fig_res_nodepred_OGBAmazon_Heter}. The amplitude of the gradient is reported according to the color scale on the right hand side.}
	\label{fig_res_vecfield_OGBAmazon_Heter}
\end{figure}

\begin{figure}[htb]
	\centering
	\includegraphics[width=1\columnwidth]{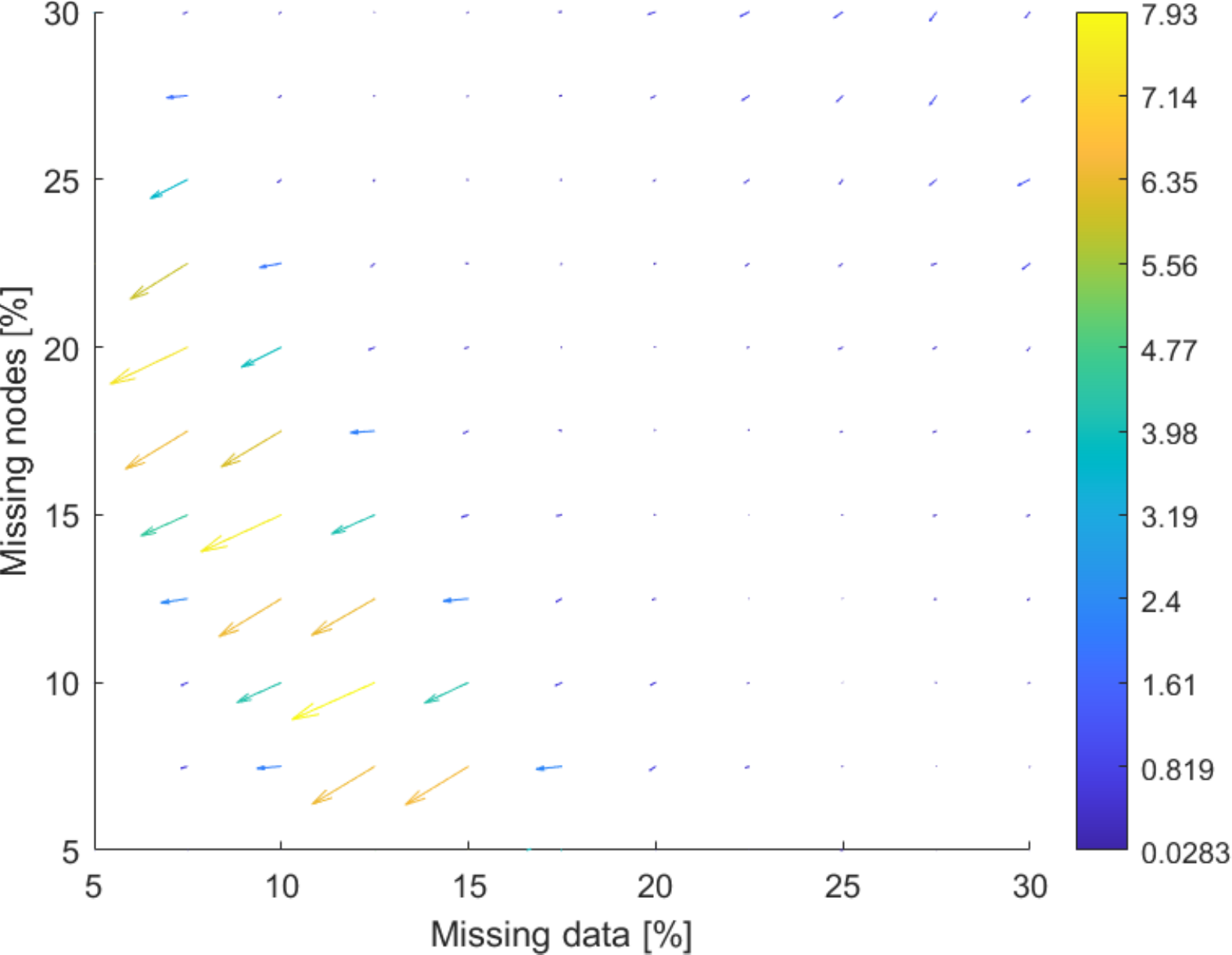} 
	\caption{Vector field representation of the ascending gradient for the average accuracy results reported in Fig. \ref{fig_res_nodepred_OGBAmazon_Symmetric}. The amplitude of the gradient is reported according to the color scale on the right hand side.}
	\label{fig_res_vecfield_OGBAmazon_Symmetric}
\end{figure}

\begin{figure}[htb]
	\centering
	\includegraphics[width=1\columnwidth]{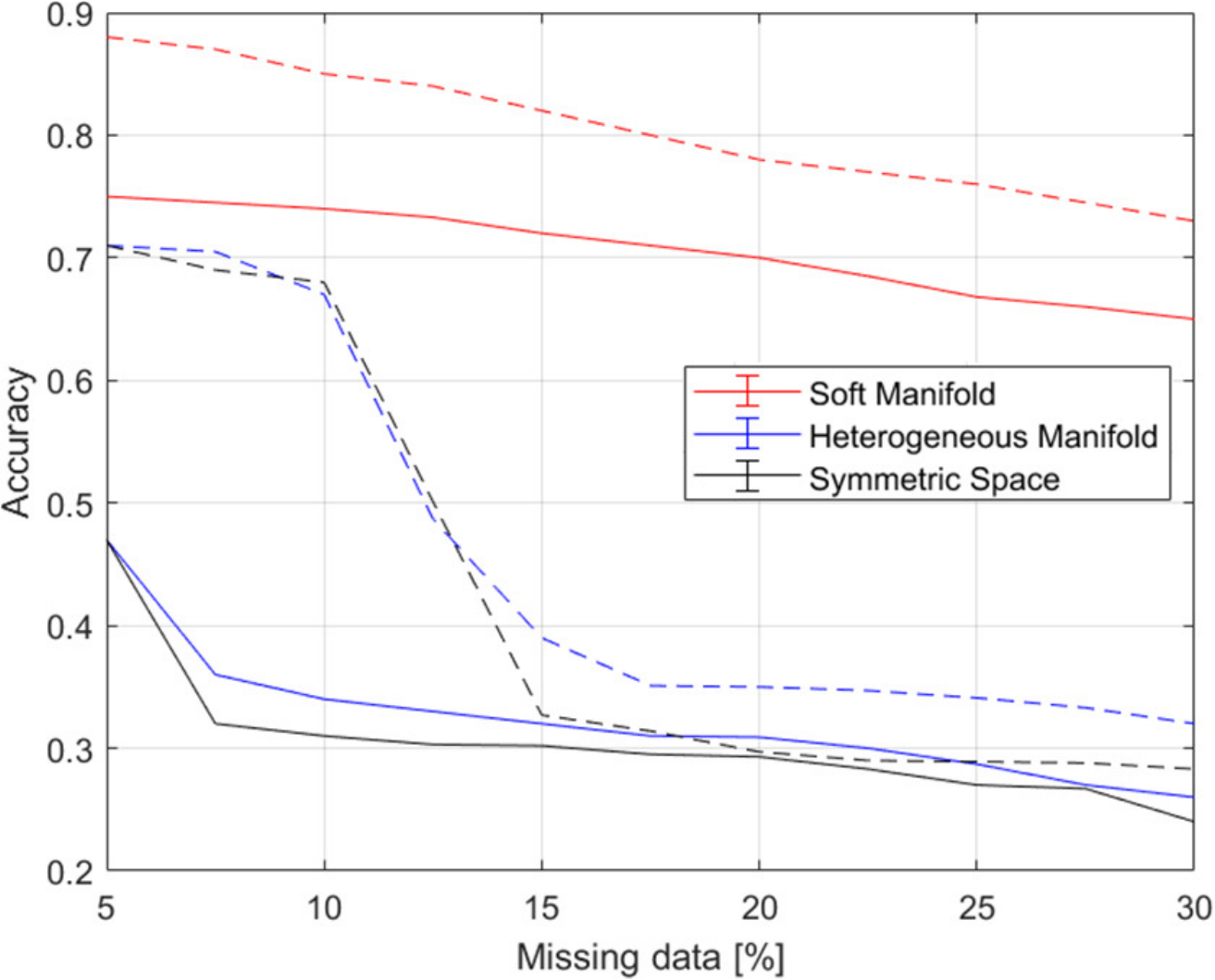} 
	\caption{Accuracy trends obtained when performing node prediction on the product co-purchasing dataset in Section \ref{sec_exp_OGBAmazon} the by means of Soft Manifold (red lines), Heterogeneous Manifold (blue lines), and Symmetric Space (black lines) with 10$\%$ and 25$\%$ of missing nodes (dashed and solid lines, respectively), as a function of the percentage of missing data. These results are retrieved from Fig. \ref{fig_res_nodepred_OGBAmazon_Soft}-\ref{fig_res_nodepred_OGBAmazon_Symmetric}. }
	\label{fig_res_transect_OGB}
\end{figure}

Other properties of the results displayed in Figures \ref{fig_res_nodepred_OGBAmazon_Soft}-\ref{fig_res_nodepred_RS_Symmetric} further support this statement. 
In particular, it is possible to observe that the slope of the performance decrease due to the increase of missing nodes and missing data seems quite uniform in Figures \ref{fig_res_nodepred_OGBAmazon_Soft} and \ref{fig_res_nodepred_RS_Soft}. 
On the other hand, the results achieved when considering the methods in \cite{HeterManifold} and \cite{GraphEmbed14} show steeper performance decrease slopes, and an apparent skewness of these performance decrease trends towards the increase of missing data for each value of missing nodes to be predicted. 
In other terms, when using these methods, the accuracy performance degrades faster than when using soft manifolds. 

\begin{figure}[htb]
	\centering
	\includegraphics[width=1\columnwidth]{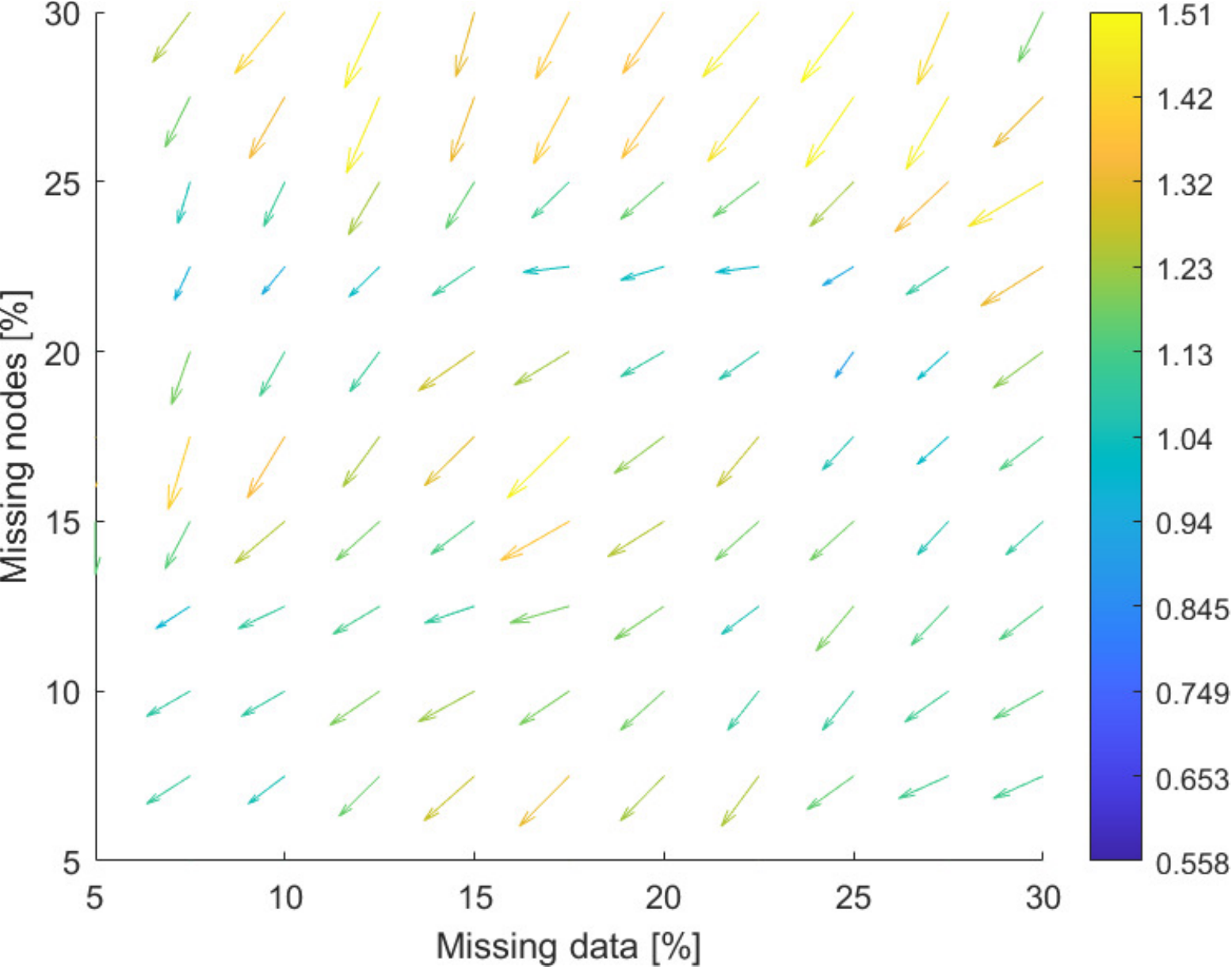} 
	\caption{Vector field representation of the ascending gradient for the average accuracy results reported in Fig. \ref{fig_res_nodepred_RS_Soft}. The amplitude of the gradient is reported according to the color scale on the right hand side.}
	\label{fig_res_vecfield_RS_Soft}
\end{figure}

\begin{figure}[htb]
	\centering
	\includegraphics[width=1\columnwidth]{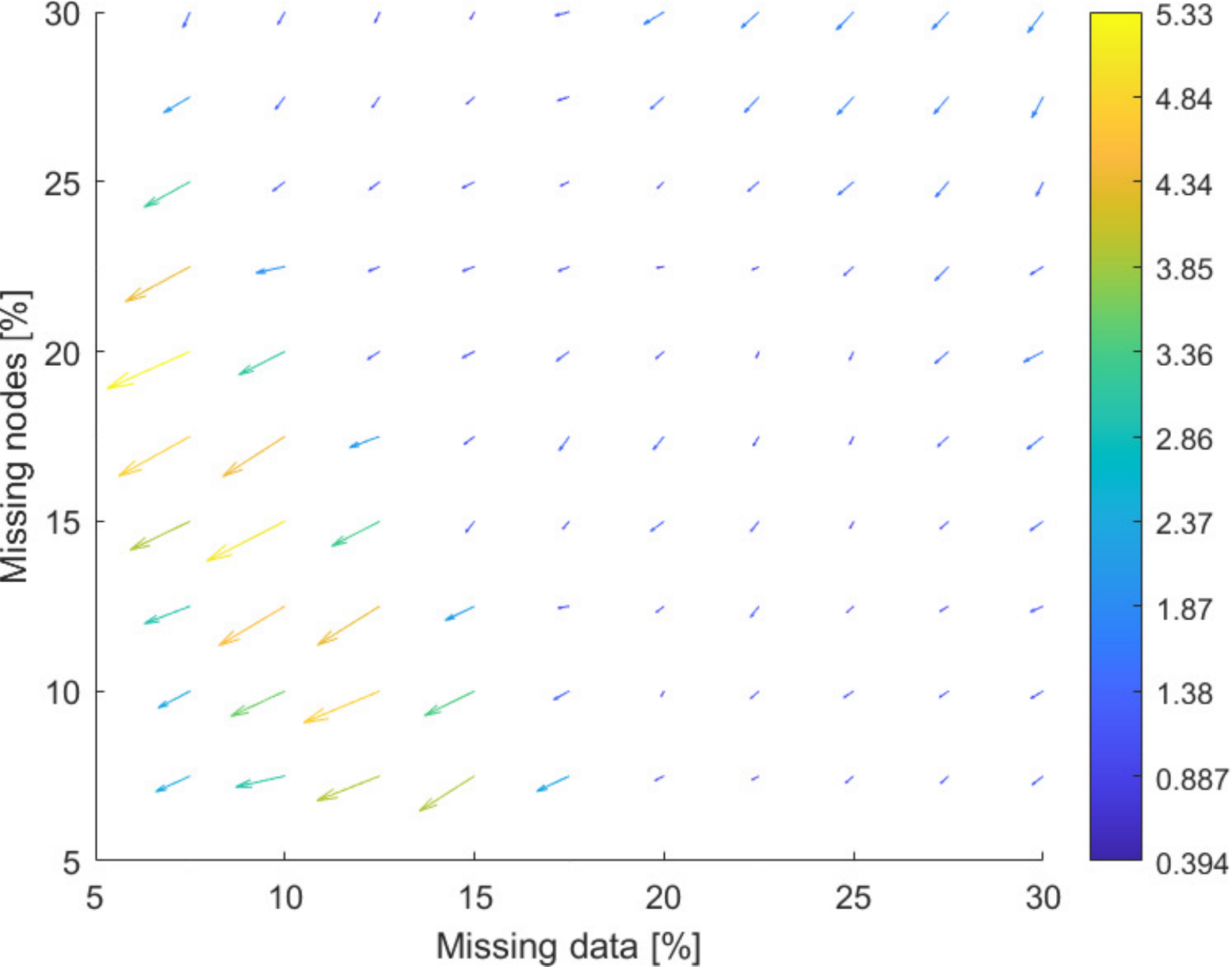} 
	\caption{Vector field representation of the ascending gradient for the average accuracy results reported in Fig. \ref{fig_res_nodepred_RS_Heter}. The amplitude of the gradient is reported according to the color scale on the right hand side.}
	\label{fig_res_vecfield_RS_Heter}
\end{figure}

\begin{figure}[htb]
	\centering
	\includegraphics[width=1\columnwidth]{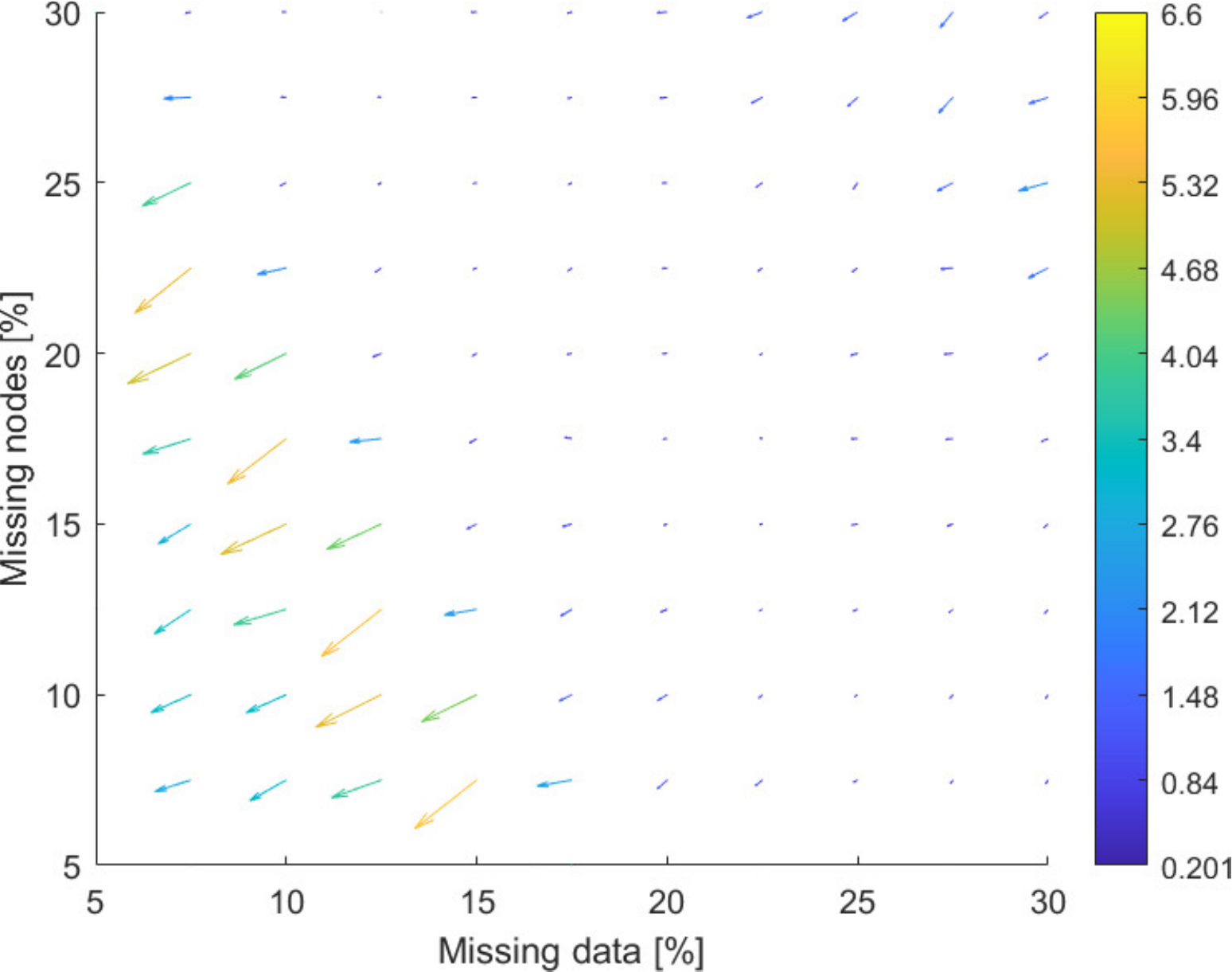} 
	\caption{Vector field representation of the ascending gradient for the average accuracy results reported in Fig. \ref{fig_res_nodepred_RS_Symmetric}. The amplitude of the gradient is reported according to the color scale on the right hand side.}
	\label{fig_res_vecfield_RS_Symmetric}
\end{figure}

Also, as the quantity of missing nodes increases, it takes a lower value of missing data to jeopardize the ability of the node prediction procedure to adequately track the mapping of the data points onto the manifold, and then provide useful information for the node prediction. 
This outcome is further emphasized by the vector fields that illustrate the direction and amplitude of the ascending gradient for the aforesaid accuracy surfaces. 
Specifically, Fig. \ref{fig_res_vecfield_OGBAmazon_Soft}-\ref{fig_res_vecfield_OGBAmazon_Symmetric} display the vector fields for the accuracy results in Fig. \ref{fig_res_nodepred_OGBAmazon_Soft}-\ref{fig_res_nodepred_OGBAmazon_Symmetric}, respectively, whilst the vector fields for Fig. \ref{fig_res_nodepred_RS_Soft}-\ref{fig_res_nodepred_RS_Symmetric} are reported in Fig. \ref{fig_res_vecfield_RS_Soft}-\ref{fig_res_vecfield_RS_Symmetric}, respectively. 

\begin{figure}[htb]
	\centering
	\includegraphics[width=1\columnwidth]{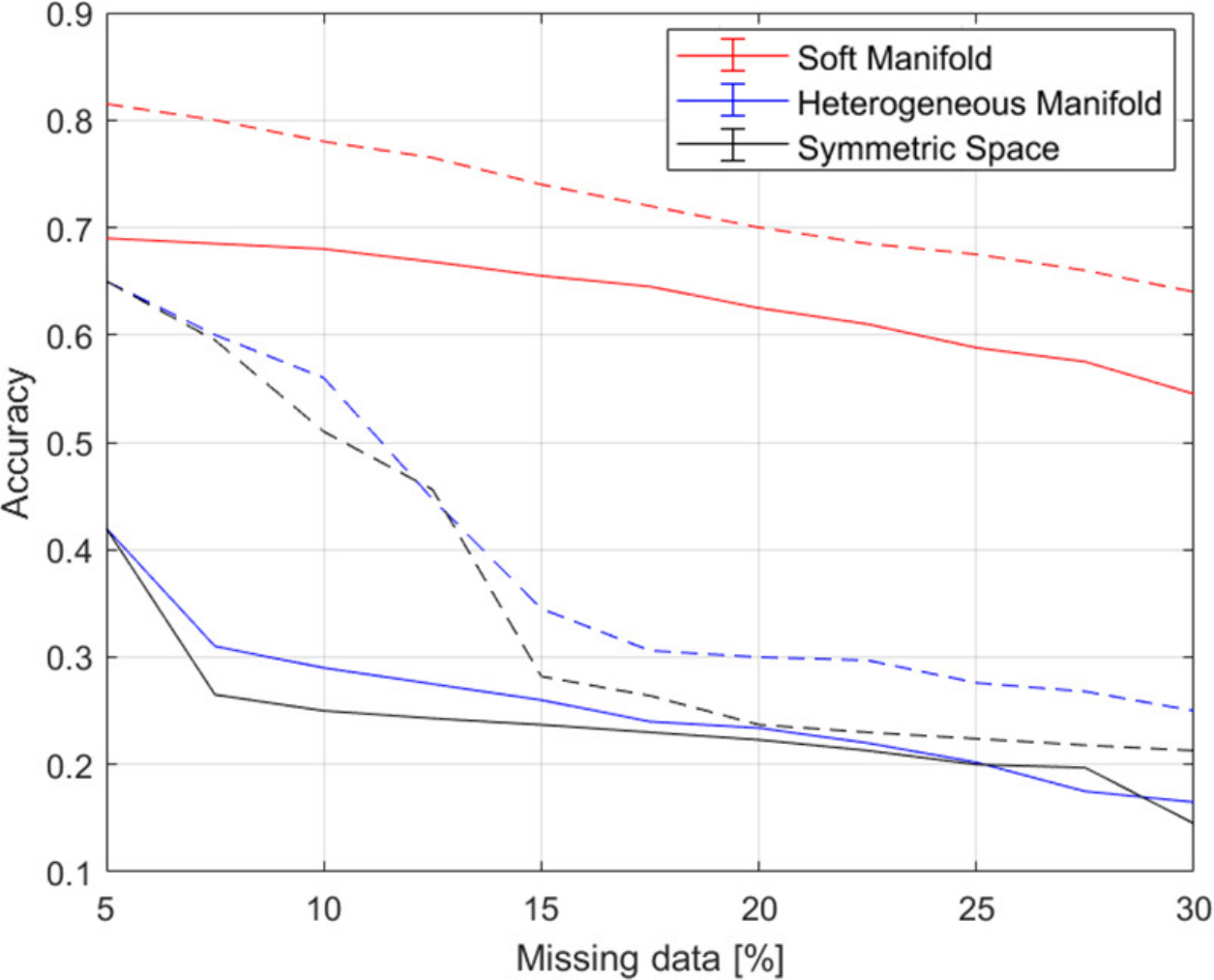} 
	\caption{Accuracy trends obtained when performing node prediction  over the remote sensing dataset in Section \ref{sec_exp_RS} by means of Soft Manifold (red lines), Heterogeneous Manifold (blue lines), and Symmetric Space (black lines) with 10$\%$ and 25$\%$ of missing nodes (dashed and solid lines, respectively), as a function of the percentage of missing data. These results are retrieved from Fig. \ref{fig_res_nodepred_RS_Soft}-\ref{fig_res_nodepred_RS_Symmetric}. }
	\label{fig_res_transect_RS}
\end{figure}

Moreover, we report in Fig. \ref{fig_res_transect_OGB} and \ref{fig_res_transect_RS} the accuracy trends that have been achieved when considering the product co-purchasing dataset in Section \ref{sec_exp_OGBAmazon} and the remote sensing dataset in Section \ref{sec_exp_RS}, respectively. 
In particular, these Figures display the outcomes obtained by 
Soft Manifold (red lines), Heterogeneous Manifold (blue lines), and Symmetric Space (black lines) with 10$\%$ and 25$\%$ of missing nodes (dashed and solid lines, respectively), as a function of the percentage of missing data. It is hence possible to appreciate that the increase of missing data in the node prediction systems strongly affects the accuracy outcomes, providing a substantial degradation of the performance. This result has several implications on the quality of information extraction that these methods can provide, especially for the use of these schemes in operational scenarios.


\section{Conclusion}
\label{secconcl}

In this paper, we introduce a novel approach for embedding on continuous space of graphs with missing data. 
The proposed scheme is based on the use of a novel type of manifolds, named soft manifolds, that are instrumental to analyze and characterize datasets with missing samples and features. 
This approach is based on the assumption that tangent planes to the point living on soft manifolds would not be Euclidean flat. 
Since soft manifolds are constructed to have spherical symmetry, the distance between points on soft manifolds would be defined in terms of hypocycloids. 
This mathematical structure allows us to define information propagation in continuous spaces in terms of fluid diffusion mechanism, which proved to be instrumental for the robust processing of datasets with variable reliability. 
To test the actual impact of the proposed method, we conducted experiments on two widely used datasets. 
The results we achieved in terms of graph embedding and node prediction show how the proposed method greatly outperforms state-of-the-art methods. 
Future works will be devoted to investigate the capacity and limits of soft manifolds for diverse tasks in automatic learning for modern data analysis, such as inference, assessment, uncertainty propagation and prediction.

\section{Acknowledgements}

This work is funded in part by Centre for Integrated Remote
Sensing and Forecasting for Arctic Operations (CIRFA) and
the Research Council of Norway (RCN Grant no. 237906), the Visual Intelligence Centre for
Research-based Innovation funded by the Research Council of
Norway (RCN Grant no. 309439), and the NATALIE project funded by the European Union Horizon Europe Climate research and innovation program under grant agreement nr. 101112859. 

\appendices

\section{Elements of fluid diffusion for data analysis}
\label{app_meth_fluid}

Let $\textbf{X} = \{ \textbf{x}_i\}_{i=1, \ldots, N}$, $\textbf{x}_i \in \mathbb{R}^n$ be the considered dataset consisting of $N$ samples characterized by $n$ features. 
It is possible to identify the characteristics of the data and the interactions across data points by considering the acquired records as results of a dynamical process at steady state \cite{fluid_diffusion1,fluid_diffusion3,fluid_arxiv}. 
In particular, it is useful to  combine the main properties of random walks and spectral analysis in order to find relevant structures in the datasets under exam. 

It is possible indeed to interpret the similarity between samples by 
using the eigenfunctions of a Markov matrix defining a random walk on the data  \cite{fluid_diffusion2,fluid_diffusion3}. 
In fact, the distances between data points can be associated with transition probabilities associated with the considered  
diffusion processes \cite{fluid_diffusion1,fluid_diffusion3}.  
Furthermore, the higher order moments of the Markov matrix provide information on the geometrical characteristics of the dataset \cite{COUILLET16}. 

Hence, when the \textbf{heat diffusion} mechanism is considered, the given dataset is $\textbf{X}$ is assumed to be sampled from the following dynamical system in
equilibrium \cite{fluid_diffusion1,fluid_diffusion3}: 

\begin{equation}
	\dot{\textbf{x}} = - \nabla U(\textbf{x}) + \sqrt{2}\dot{\textbf{w}},
	\label{eq_dynsyst_heat}
\end{equation}

\noindent where $\dot{\textbf{x}}$ and $\dot{\textbf{w}}$ identify the derivatives with respect to $t$ of $\textbf{x}$ and $\textbf{w}$, respectively.  
Moreover, $U$ is the free energy at
$\textbf{x}$ (which can be also called the potential at $\textbf{x}$), and $\textbf{w}$(t) is
an $n$-dimensional Brownian motion process. 
Consequently, the transition probabilities used to compute the distances between data points are derived by solving the associated forward Fokker-Planck equation for heat diffusion process which can be written for the density $p(\textbf{x}(t+\epsilon) = \textbf{x}_j|\textbf{x}(t)=\textbf{x}_i)$ as follows \cite{fluid_diffusion1}:

\begin{equation}
	\frac{\partial p}{\partial t} = \nabla \cdot (\nabla p + p\nabla U(\textbf{x})),
	\label{eq_FokkerPlanck_heat}
\end{equation}

\noindent where $\nabla = [\frac{\partial}{\partial x_i}]_{i=1, \ldots, n}$. 
The solution for the transition probability associated with the $i$-th and $j$-th data points under the heat diffusion assumption can be written as follows \cite{Kondor02,fluid_kernel1}:

\begin{equation}
	p(\textbf{x}(t+\epsilon) = \textbf{x}_j|\textbf{x}(t)=\textbf{x}_i) = p_{ij} \propto \exp \left[  - \frac{||\textbf{x}_i - \textbf{x}_j ||_2}{2 \sigma} \right],
\end{equation}

\noindent where $\sigma$ is the variance associated with the considered dataset \cite{fluid_diffusion1,fluid_diffusion3}. 

Considering the complexity of modern data analysis, summarized in Section \ref{sec_back} in the data characteristics \textbf{C1}-\textbf{C3}, the heat diffusion model might is inadequate to support the reliable information extraction task \cite{GEOMDL_limits_2,GEOMDL_limits_8,fluid_arxiv,fluid_ICASSP23}.
To overcome these issues, the diffusion model should show an higher degree of flexibility to capture the full complexity of the datasets. 
In \cite{fluid_arxiv}, a model based on \textbf{fluid diffusion} mechanism is used to retrieve a more thorough and complete understanding of the datasets than the heat diffusion one is proposed. 
Hence, the system in (\ref{eq_dynsyst_heat}) should be replaced by a more complex stochastic differential model, such as follows:

\begin{equation}
	\dot{\textbf{x}} = \textbf{a}(\textbf{x}) + \textbf{B}(\textbf{x})\dot{\textbf{w}},
	\label{eq_dynsyst_fluid}
\end{equation}

\noindent where $\textbf{w}(t)$ is a $N$-dimensional Wiener process, 
$\textbf{a}(\textbf{x})$ is a length-$n$ vector, whilst 
$\textbf{B}(\textbf{x})$ identifies a $n \times N$ matrix  \cite{fluid_FokkerPlanck3,fluid_FokkerPlanck5,fluid_FokkerPlanck6,fluid_FokkerPlanck4}.  
Further, the \textbf{a} term is typically written as follows: 

\begin{equation}
	\textbf{a}(\textbf{x}) = \textbf{v}(\textbf{x}, {\cal K}) + \nabla \cdot \tilde{\textbf{B}}(\textbf{x}).  
	\label{eq_flowrate}
\end{equation}

The quantities in (\ref{eq_dynsyst_fluid}) and (\ref{eq_flowrate}) can be described in terms of fluid dynamics properties, i.e.:

\begin{itemize}
	\item  the $\textbf{B}(\textbf{x})$ matrix summarizes the rate by which the diffusion can take place across the features of the considered system \cite{fluid_FokkerPlanck5,fluid_FokkerPlanck2}; 
	\item \textbf{a} regulates the \textit{flow rate}, i.e., the velocity by which the diffusion can take place from one node to another in the system \cite{fluid_FokkerPlanck5,fluid_FokkerPlanck2}; 
	\item the \textit{conductivity} properties (summarized by a $N \times N \times n$ tensor $\cal K$) model the ease with which the fluid diffusion can take place from one node to another in the system \cite{fluid_FokkerPlanck2,fluid_FokkerPlanck1,fluid_FokkerPlanck3};
	\item the matrix $\tilde{\textbf{B}}(\textbf{x}) = \frac{1}{2} \textbf{B}(\textbf{x})\textbf{B}^T(\textbf{x})$ models the ability of each feature to permit diffusion across the nodes in the diffusion system;
	\item \textbf{v} models the transport velocity of information between data points. The velocity between sample \textit{i} and sample \textit{j} is typically computed as $v_{ij} = - || {\cal K}_{ij:}^T \odot (\textbf{x}_i - \textbf{x}_j) ||_2 \in [0,1]$, where ${\cal K}_{ij:}$ is the length-$n$ row vector collecting the third dimension elements of the conductivity tensor $\cal K$ on the $(i,j)$ coordinates and $\odot$ is the Hadamard product  \cite{fluid_FokkerPlanck1,fluid_FokkerPlanck2,fluid_FokkerPlanck3,fluid_FokkerPlanck4, fluid_FokkerPlanck5, fluid_FokkerPlanck6, Fluid_TDRW1, Fluid_TDRW2}.  
\end{itemize}

The diffusion equation associated with the system in (\ref{eq_dynsyst_fluid}), and used to retrieve the transition probabilities between data points (and their distances hence), is the Fokker-Planck equation for fluid diffusion in porous media, and can be written as follows 
\cite{fluid_FokkerPlanck1,fluid_FokkerPlanck2}: 

\begin{eqnarray}
	\frac{\partial p(\textbf{x},t)}{\partial t} = & -&  \nabla \cdot \left \{ \left[\textbf{a}(\textbf{x}) - \nabla \tilde{\textbf{B}}(\textbf{x})\right]p(\textbf{x},t) \right \} \nonumber \\
	&+& \nabla \cdot \tilde{\textbf{B}}(\textbf{x})\nabla p(\textbf{x},t).
	\label{eq_FokkerPlanck_fluid_app}
\end{eqnarray}

With this in mind, the transition probability for sample \textit{i} and \textit{j} $p(\textbf{x}(t+\epsilon) = \textbf{x}_j|\textbf{x}(t)=\textbf{x}_i) = p_{ij}$ that solves (\ref{eq_FokkerPlanck_fluid_app}) can be written as follows \cite{fluid_arxiv,Fluid_TDRW1,Fluid_TDRW2}: 

\begin{equation}
	p_{ij} = \frac{|v_{+}^\dagger| \exp\left[v_{+}^\dagger \right] \csch \left[|v_{+}^\dagger|\right]}{\sum_{u \in \{+,-\}} |v_{u}^\dagger| \exp\left[\tilde{u}\cdot v_{u}^\dagger \right] \csch \left[|v_{u}^\dagger|\right]},
	\label{eq_fluid_trans_prob_pij}
\end{equation}

\noindent where $\csch[z] = 1/\sinh[z] = 2/(\exp[z] - \exp[-z])$. Moreover, 
$\tilde{u}$ is set to 1 when $u=+$, whilst $\tilde{u}=-1$ if $u$ is -. Finally, $v_{\pm}^\dagger = v_{\pm}/2\tilde{B}_\pm$, being: 

\begin{eqnarray}
	v_+ & = & v_{ij}, \\
	v_- & = & \sum_{m \in {\cal N}(i) \setminus j} \frac{v_{im}}{|{\cal N}(i)|- 1}, \nonumber \\
	\tilde{B}_+ & = & \tilde{B}_{ij}, \nonumber \\
	\tilde{B}_- & = & \sum_{m \in {\cal N}(i) \setminus j} \frac{\tilde{B}_{im}}{|{\cal N}(i)|- 1}, \nonumber
	\label{eq_v_B_pm}
\end{eqnarray}

\noindent where ${\cal N}(i)$ identifies the neighborhood of sample $i$, i.e., the set of data points adjacent to sample $i$.  

\section{Invariance and differentiable continuous maps}
\label{app_map}

Let us define $\Sigma({\cal L}_\textbf{u}) = \{ \lambda_k \}_{k \in \mathbb{N}_0}$ the spectrum of the eigenvalues of the linear operator ${\cal L}_\textbf{u} = -\Delta + \textbf{u} \cdot \nabla$ on the Hilbert space ${\cal H}^2 = {\cal H}^2(d \mu)$, being $d \mu = r d\textbf{x}$, and $\lambda_k \leq \lambda_{k+1}$ $\forall k \in \mathbb{N}_0$. 
For any $K \in \mathbb{N}_0$, we can consider the subspace ${\cal T}$ of ${\cal H}_y^2$ that is spanned by the first $K$ eigenvalues, and define ${\cal T}_O$ its orthogonal complement, thus ${\cal H}_y^2 = {\cal T} \oplus {\cal T}_O$. 
At this point, we can define the semiflow associated with (\ref{eq_FokkerPlanck_fluid_porousmedium2}) as $\Phi^t_{\tau, \upsilon}: {\cal H}_y^2 \rightarrow {\cal H}_y^2$ s.t. $\Phi^t_{\tau, \upsilon}(y^*)  = y$ if $y$ is the solution of (\ref{eq_FokkerPlanck_fluid_porousmedium2}) when the initial conditions are set to $y^* \in {\cal B}_{\tau,\upsilon} = \{ y \in C^{0,1}: ||y||_{L^\infty} \leq \tau, ||y||_{\Lip} \leq \upsilon \}$, as previously mentioned. 

Finally, let us assume there exist 
a function $\theta: {\cal T} \rightarrow {\cal T}_O$ that is a \textit{differentiable map} from ${\cal B}_{\tau, \upsilon}$ to $C^{0,1}$, and $\theta(0)=0$, $d_\texttt{F}\theta(0) = 0$, and $||\theta(\eta_{\cal T})||_{C^{0,1}} \lesssim ||\eta_{\cal T}||^2_{C^{0,1}}$ $\forall \eta_{\cal T} \in {\cal T} \cap {\cal B}_{\tau,\upsilon}$, where $d_\texttt{F}(\cdot)$ is the Fresnel derivative. 
Also, let us consider the 
manifolds $\Omega_{\cal T} = \{ \eta_{\cal T} + \theta(\eta_{\cal T}): \eta_{\cal T} \in {\cal T} \}$ and $\Omega_{I} = \Omega_{\cal T} \cap {\cal B}_{\tau, \upsilon}$
for $\tau, \upsilon >0$. 
Then, we can state that \cite{InvariantManifolds}:
\begin{itemize}
	\item the manifold $\Omega_{I}$ (local center manifold) is tangent at the origin to the eigenspace ${\cal T}$; 
	\item $\Omega_{I}$ is locally flow invariant
	under the semi-flow $\Phi^t_{\tau, \upsilon}$
	associated with (\ref{eq_FokkerPlanck_fluid_porousmedium2}) 
	for nonlinear perturbations.  
\end{itemize}

Then, for $(y^*,t) \in {\cal H}_y^2 \times [0, +\infty[$   we can define the map $(y^*,t) \mapsto \Phi^t_{\tau, \upsilon}(y^*) \in {\cal H}_y^2$ that is \textit{continuous and differentiable} \cite{InvariantManifolds}. 

\bibliographystyle{unsrt}\scriptsize	
\bibliography{BIGDATArefs_2, BigDATArefs_old}

\end{document}